\documentclass[11pt,a4paper]{article}
\usepackage[dvipsnames,table,xcdraw]{xcolor}
\usepackage[hyperref]{acl2021}
\usepackage{times}
\usepackage{latexsym}
\usepackage{epsfig}
\usepackage{graphicx}
\usepackage{amsmath}
\usepackage{amssymb}
\usepackage{array, multirow}
\usepackage{amsfonts}
\usepackage{mathtools}
\usepackage{rotating} 
\usepackage{rotfloat}
\usepackage{makecell}

\usepackage{xcolor}
\usepackage{xspace}
\usepackage{booktabs} % To thicken table lines
\usepackage{enumitem}
\usepackage{flushend}
\usepackage{pifont}% http://ctan.org/pkg/pifont
\usepackage{siunitx}

\newcommand{\cmark}{\textcolor{NavyBlue}{\ding{51}}}%
\newcommand{\xmark}{\textcolor{BrickRed}{\ding{55}}}%
\DeclareMathOperator*{\argmax}{arg\,max}

\makeatletter
\DeclareRobustCommand\onedot{\futurelet\@let@token\@onedot}
\def\@onedot{\ifx\@let@token.\else.\null\fi\xspace}

\usepackage{microtype}

\aclfinalcopy % Uncomment this line for the final submission
\def\aclpaperid{1649} %  Enter the acl Paper ID here

\title{DVD: A \underline{D}iagnostic Dataset for Multi-step Reasoning \\
in \underline{V}ideo Grounded \underline{D}ialogue}

\author{Hung Le$^{\ddag}{^\S}$\thanks{\hspace{0.1cm} Work done when HL was a research intern at Facebook.}, 
Chinnadhurai Sankar$^{\dag}$, 
Seungwhan Moon$^{\dag}$,
Ahmad Beirami$^{\dag}$, \\
\textbf{Alborz Geramifard}$^{\dag}$,
\textbf{Satwik Kottur}$^{\dag}$ \\
  $^{\dag}$Facebook\\
  \texttt{\{chinnadhurai, shanemoon, beirami, alborzg, skottur\}@fb.com}\\
 $^{\ddag}$Singapore Management University\\
 $^\S$Institute for Infocomm Research, A*STAR \\
  \texttt{hungle.2018@smu.edu.sg}}

\date{}

\begin{document}
\maketitle
\begin{abstract}
A video-grounded dialogue system is required to understand both dialogue, which contains
semantic dependencies from turn to turn, and video, which contains visual cues of spatial and
temporal scene variations.
Building such dialogue systems is a challenging problem, involving various reasoning types on both visual and language inputs.
Existing benchmarks do not have enough annotations to thoroughly analyze dialogue systems and understand their capabilities and limitations in isolation. 
These benchmarks are also not explicitly designed to minimise biases that models can exploit without actual reasoning. 
To address these limitations, in this paper, we present \textit{\textbf{DVD}, a \textbf{D}iagnostic Dataset for \textbf{V}ideo-grounded \textbf{D}ialogues}.
The dataset is designed to contain minimal biases and has detailed annotations for the different
types of reasoning over the spatio-temporal space of video. 
Dialogues are synthesized over multiple question turns, each of which is injected with a set of cross-turn semantic relationships.  
We use DVD to analyze existing approaches, providing interesting insights into their abilities and limitations.
In total, DVD is built from $11k$ CATER synthetic videos and contains $10$ instances of $10$-round dialogues for each video, resulting in more than $100k$ dialogues and $1M$ question-answer pairs.
Our code and dataset are publicly available\footnote{\url{github.com/facebookresearch/DVDialogues}}.
\end{abstract}

\section{Introduction} 
\begin{figure}[htbp]
	\centering
	\resizebox{1.0\columnwidth}{!} {
	\includegraphics{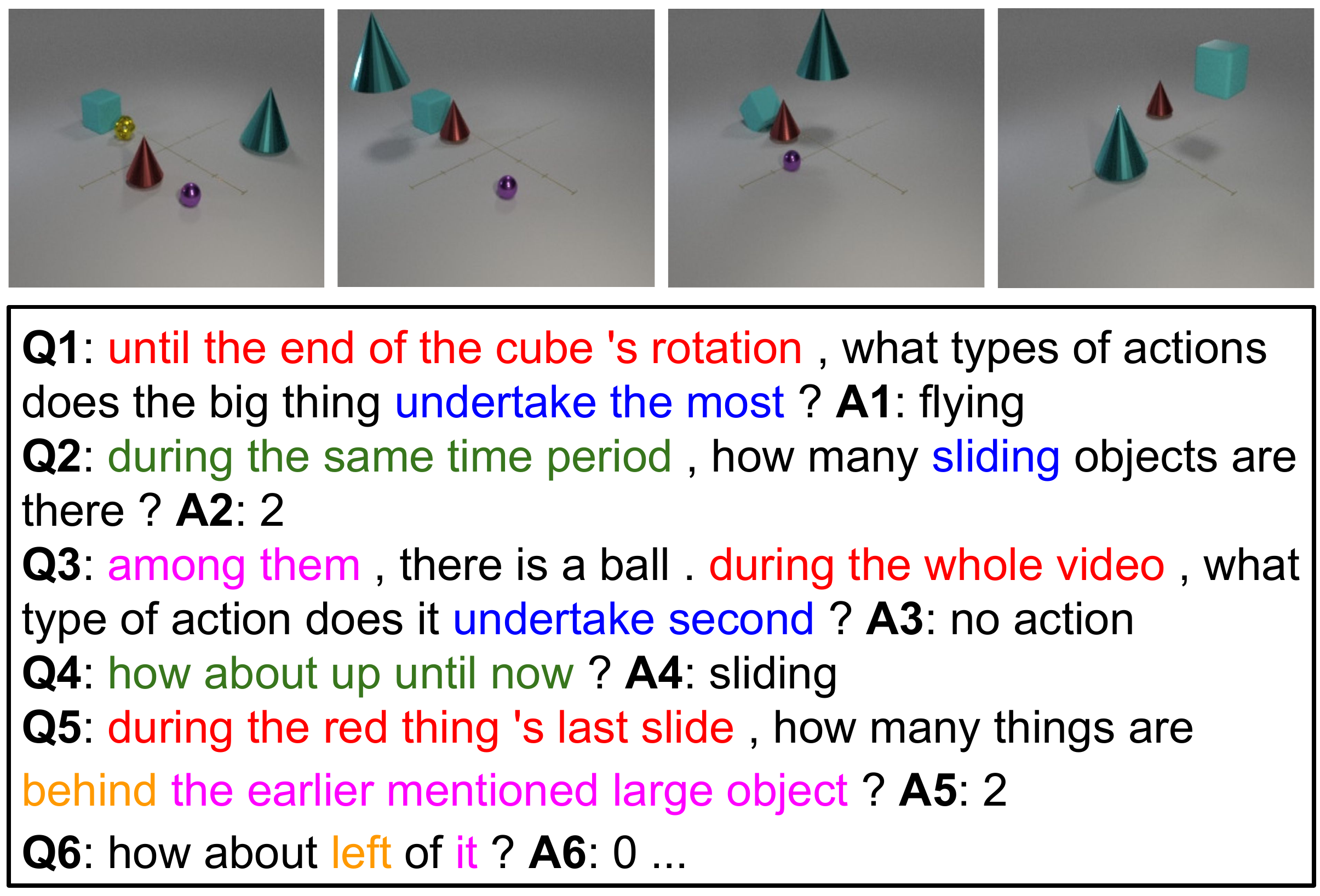}
	}
	\caption{
	\textbf{Example DVD dialogue:} We demonstrate an example dialogue in DVD that tests various aspects, including \textcolor{blue}{action recognition},
	\textcolor{red}{temporal reasoning},
	\textcolor{Orange}{spatial reasoning},
	\textcolor{ForestGreen}{video interval tracking}, and 
	\textcolor{magenta}{dialogue object tracking}.
	%As illustrated, each of questions from the second dialogue turns has some form of cross-turn relations to past dialogue turns. 
	%Rather than a fixed video content, the visual input is simulated from a continuous video stream.
	$\mathrm{Q}_i/\mathrm{A}_i$: question/answer of turn $i$. 
	%$\mathrm{EOV}_j$: end of video input $\mathrm{V}_j$. 
	}
	\label{fig:example_dial}
	\vspace{-0.2in}
\end{figure}

Research in visual question answering (VQA) 
%is a popular
%\sk{SK: Calling 5 years long-studied might be a stretch. Do you want to say popular instead?}
%line of research that 
aims to develop intelligent systems that can reason and answer questions about visual information.
Earlier datasets have been introduced to study this problem, focusing on images as the visual input \cite{antol2015vqa, gao2015you, malinowski2014multi, zhu2016visual7w}
%\sk{
%SK: Say that these works focus on images only, so that the next sentence flows better
%}
Recently, many QA benchmarks have been proposed to extend the visual information from the image to video domain \cite{jang2017tgif, lei-etal-2018-tvqa, zadeh2019social}.
While image QA problems require a system to learn cross-modality interaction, video QA problems go beyond and capture visual information with temporal variance. 
%\sk{
%Since we did not talk about challenges in VQA, 'increases' might not be impactful.
%We should mention challenges in VQA and say that video QA goes beyond and captures another dimension.
%}
%\sk{
%SK: Video QA and Visual Dialog are kind of contemporary (2017) according to our citations.
%More recently in that case might be inconsistent. Should we talk about these are
%orthogonal extensions from VQA? Also, cite guesswhich paper for Visual Dialog
%}
%Compared to traditional VQA tasks, the complexity of the video QA problems increases by the additional temporal variance across video frames. 
%Correctly answering questions about the content of videos requires different types of perceptual abilities, such as recognizing moving objects, including their locations and actions.

As an orthogonal extension from VQA problems, another line of research investigates image/video QA in a dialogue setting \cite{das2017visual, seo2017visual, de2017guesswhat, visdial_eval, alamri2019audio}.
In this problem, questions about a given video or image are positioned in a multi-turn dialogue. 
In each dialogue turn, a question usually exhibits different types of cross-turn relations to other questions in prior dialogue turns, such as object co-reference and topic alignment.
In this work, we investigate the problem of multi-turn video question answering (QA), also known as video-grounded dialogue. 

Numerous approaches to video-grounded dialogue have shown remarkable performance in building intelligent multimodal systems \cite{hori2019avsd, schwartz2019factor, le-etal-2019-multimodal, li2020bridging, le-etal-2020-bist}.
However, most of these methods exhibit marginal performance gain, and our ability to understand their limitations is impeded by the complexity of the task.
Existing benchmarks are not designed with enough information to 
determine whether current approaches are capable of sophisticated reasoning and not just exploiting biases, which has been a common concern in vision-language systems \cite{agrawal-etal-2016-analyzing, goyal2017making, qi2020two}. 

To address the limitations of existing benchmarks and analyze dialogue systems more efficiently, we propose \textit{\textbf{DVD}, a \textbf{D}iagnostic Dataset for \textbf{V}ideo-grounded \textbf{D}ialogues}.
We demonstrate an example dialogue in DVD in Figure \ref{fig:example_dial}. 
%We utilized about $11k$ synthetic videos from CATER \cite{Girdhar2020CATER} and built a benchmark of over $100k$ dialogues and $1M$ questions. 
%We built our benchmark on top of a challenging video dataset, CATER \cite{Girdhar2020CATER}.
%CATER dataset contains videos of multiple objects arranged in a 3D environment with high variance of object appearance, locations, and actions. 
%Moreover, the information in each video is not affected by external information sources, such as commonsense knowledge, making it ideal for learning visual reasoning in dialogues. 
From scene graphs and object action annotation of a CATER video \cite{Girdhar2020CATER}, we simulate questions based on reasoning structures, also known as functional programs in CLEVR \cite{johnson2017clevr}.
Compared to CLEVR, we introduced 17 novel functional modules, designed for video and dialogue input components. 
As illustrated in Figure \ref{fig:example_dial}, at each dialogue turn, a DVD question tests dialogue systems to perform different types of reasoning on videos, such as action recognition and spatio-temporal reasoning. 
Across turns, we generate questions to be related to each other by incorporating different types of semantic relationships, including: 
(1) temporal relation, which requires a system to learn to localize different temporal segments of the video from turn to turn; 
(2) object reference, which requires a system to resolve visual objects mentioned throughout the dialogue history in either short-term references (pronouns) or long-term references (e.g. ``the earlier mentioned large object'');
and (3) topic transfer, which requires a system to maintain a memory of the last question turn to solve the question in the current turn.

%Multiple factors can affect the performance of a system.
%For instance, a dialogue agent answers incorrectly if it is unable to decode dialogue context and derive contextualized objects, such as ``the earlier mentioned large object'' and ``it'' in Questions $5$ and $6$ of Figure \ref{fig:example_dial}.  
%Moreover, a video-grounded dialogue does not just focus on a specific timestamp or video segment.
%Instead, different video segments are mentioned from turn to turn and a system is required to locate the right temporal information (See Figure \ref{fig:example_dial}). 

%Specifically, we proposed two sub-tasks relevant to video-grounded dialogue problems, \textit{video interval tracking (\textbf{VIT})}, and \textit{dialogue object tracking  (\textbf{DOT})}.
%The VIT task requires a dialogue system to identify which video segment each dialogue turn is referring to. In each turn, the question is generated such that the corresponding video segment is either independent or related to another segment in prior dialogue turns. 
%This task is an extension of prior research in temporal localization through text, or text-to-clip \cite{anne2017localizing} but is designed in a multi-turn setting. 
%DOT has the dialogue state tracking (DST) nature, which requires a system to track information slots in task-oriented dialogues \cite{mrksic-etal-2017-neural}. 
%While in task-oriented dialogues, the tracked slots are used to create API queries to entity databases, tracked objects in DOT is used to solve object references and locate the visual objects from videos. 

On DVD, we trained a set of baseline methods and analyzed the results by several aspects of visual and linguistic complexity (Section \ref{sec:exps}). 
We found that these methods struggle on questions requiring both video temporal and spatial localization.
They are also vulnerable to long-term reasoning in both videos and dialogues as they are not designed to track active visual objects or relevant video segments throughout dialogue context. 
We hope the DVD dataset will lead to new research avenues to develop intelligent systems capable of complex reasoning on video and dialogue medium (further discussion in the Supplementary Material). 
The DVD dataset and code will be made public.

\section{Related Work} 
%Table \ref{tab:related_work} presents a comparison between DVD and relevant benchmarks. 
We compared DVD to existing datasets from the following four angles: 
%visual-linguistic, 2) visually-grounded, 3) diagnostic, and 4) multi-step reasoning.

\noindent \textbf{1) Vision-linguistic.}
 Vision-linguistic understanding benchmarks have been proposed, including captioning \cite{farhadi2010every, lin2014microsoft, rohrbach2015dataset}, phrase grounding or object reference \cite{kazemzadeh2014referitgame, plummer2015flickr30k}, scene graph learning \cite{krishna2017visual}, and text-to-clip \cite{anne2017localizing}. Our benchmark, DVD, is more related to VQA in which a visual input is given and a system is required to answer a question about this input \cite{antol2015vqa, zhu2016visual7w, jang2017tgif, lei-etal-2018-tvqa}.
Another related line of research is the research of navigation systems in a physical environment \cite{gordon2018iqa, eqa_matterport}.
%In this task, systems are required to understand and follow an instruction to navigate in a simulated physical space. 
Compared to the prior benchmarks, one major difference of DVD is the extension of single-turn interaction to a multi-turn human-machine dialogue.
%Prior models may not perform well in a multi-turn setting which contains the additional contextual information.

\noindent \textbf{2) Visually-grounded Dialogue.} 
Extended from the vision-linguistic understanding research, this line of research focuses on answering questions sequentially positioned over multiple turns \cite{de2017guesswhat, das2017visual, visdial_eval, hori2019avsd, thomason:corl19}.
%This task simulates a scenarios where human and machine exchanges information in a natural dialogue.
A system has to understand the dialogue context and resolve cross-turn semantic dependencies. 
However, due to the complexity of the tasks, involving cross-modality and cross-turn information, prior benchmarks are often subject to bias that models often exploit without actual reasoning \cite{qi2020two}.
%Specifically, in a video-grounded dialogue, each question turn focuses on different parts of the video.
%One limitation of current benchmarks is the lack of information to evaluate how a model localizes relevant video intervals from turn to turn.   
In this work, we design a diagnostic benchmark with minimal bias and incorporate a  set of specific reasoning requirements. 

\noindent \textbf{3) Diagnostic.}
Our work is related to MNIST Dialogue \cite{seo2017visual} and CLEVR Dialog \cite{kottur-etal-2019-clevr}.
They involve synthetic images to develop image-grounded dialogues.
Compared to them, DVD questions are extended from the image to the video domain and injected with more diverse cross-turn semantics. 
%Specifically, in DVD, we incorporated 3 types of relationships: temporal relation, object reference, and topic transfer (See Section \ref{sec:method}).
%Our benchmark contains synthetic dialogues with higher linguistic variance in questions.  
As shown in Table \ref{tab:data_stats} DVD contains a higher proportion of unique questions than related benchmarks.
DVD is also inspired by the dialogue state tracking task (DST) \cite{mrksic-etal-2017-neural, DBLP:conf/iclr/BordesBW17, DBLP:journals/corr/abs-2104-08667,moon-etal-2020-situated}.  
DST requires a system to detect all information slots mentioned in dialogue, such as restaurant name and booking date.
Instead, in DVD, for each turn, we introduce an object tracking state, defined as visual objects and their attributes mentioned in dialogue context. 
%DVD is the first diagnostic multimodal dialogue benchmark that provides such detailed annotations of dialogue states.

%\textbf{Situated systems}.
%Our benchmark is also related to previous work that situate a system in a dynamic environment.
%A prominent line of research in this area is the research of navigation agents. 
%In this line of research, a system is required to move around in a physical environment following natural instructions through either single-turn \cite{gordon2018iqa, eqa_matterport} or multi-turn interaction \cite{thomason:corl19, nguyen-daume-iii-2019-help, nguyen2019vision}.

\noindent \textbf{4) Multi-step reasoning.}
A multi-step reasoning question is typically represented by a reasoning structure, also known as functional programs.
Earlier efforts \cite{andreas2016neural, johnson2017clevr} designed questions that are expressed as elementary operation programs. 
More related to our work, \citet{song2018explore, Yi*2020CLEVRER} extended the prior work to the video domain with questions focusing on the temporal variance of video frames.
A major difference between our work and these approaches is the extension of functional programs to a dialogue task with context-based operations, such as object tracking and interval tracking. 
This extension brings a step toward more transparent dialogue systems capable of performing reasoning operations across question turns. 

%\section{Diagnostic Video-grounded Dialogues}
\section{The DVD Dataset}
\label{sec:method}

Our benchmark provides a dataset that can be used to conduct rich diagnostics to better understand the reasoning capabilities of dialogue systems. 
%Following prior benchmarks using synthetic domain data \cite{johnson2017clevr, Yi*2020CLEVRER, Baradel2020CoPhy},
%We utilize CATER \cite{Girdhar2020CATER} to automatically generate dialogues. 
%The videos have associated ground-truth object attributes, locations, and action time intervals.
%The generated dialogues contain questions that have an associated functional program structures. 
%As noted by CLEVR, these programs allow use to analyze models based on different aspects.
%In this benchmark, we focus on several types of visual and linguistic operations, such as temporal localization, object tracking, and spatio-temporal reasoning. 
Table \ref{tab:data_stats} and Figure \ref{fig:functional_program} to \ref{fig:dial_framework} give an overview of DVD. 

\begin{table}[t]
\resizebox{1.0\columnwidth}{!} {
\begin{tabular}{lllll}
\hline
Split        & \begin{tabular}[c]{@{}l@{}}\#Videos/\\ Images\end{tabular} & \#Dialogs & \#Questions & \begin{tabular}[c]{@{}l@{}}\# Unique\\ Questions\end{tabular} \\
\hline
DVD-Train        & 6,157                                                      & 61,551    & 615,510     & 360,334                                                       \\
DVD-Val          & 1,540                                                      & 15,396    & 153,960     & 99,211                                                        \\
DVD-Test         & 3,299                                                      & 32,978    & 329,780     & 200,346                                                       \\
DVD-Total        & 10,996                                                     & 109,925   & 1,099,250   & 620,739                                                       \\
\hline
CLEVR        & 100K                                                       & N/A       & 1M          & 854K                                                          \\
CLEVRER      & 20K                                                        & N/A       & 305K        & 26.4K                                                           \\
VisDial      & 123K                                                       & 123K      & 1.2M        & 380K                                                          \\
AVSD         & 11.1K                                                      & 11.1K     & 101.2K      & 59K                                                           \\
MNIST Dialog & 50K                                                        & 150K      & 1.5M        & 355                                                           \\
CLEVR Dialog & 85K                                                        & 425K      & 4.25M       & 73K               \\                              \hline             
\end{tabular}}
\caption{
\textbf{Statistics for DVD}: 
Compared to synthetic dialogue benchmarks, MNIST Dialog and CLEVR Dialog, majority of questions in DVD are unique.
Questions are generated from question templates and incorporated with various cross-turn semantics. 
}
\label{tab:data_stats}
\vspace{-0.2in}
\end{table}

\subsection{Objects, Spatial Relations, and Intervals}
\textbf{Objects.} 
Objects are identified by their attributes, including object shapes,
%(cube, sphere, cylinder, cone, and snitch), 
sizes,
%(small, large, and medium), 
materials, 
%(rubber and metal), 
and colors. 
One unique characteristic of CATER objects is that each object can move multiple times in a single video. 
From the CATER universe, we define $4$ types of object actions: ``flying'', ``rotating'', ``sliding'', and ``no action'' (object being stationary).
Another characteristic of CATER objects is that one object can be contained by another object, resulting in a visual problem called \emph{object containment}.
In our experiments, current dialogue systems are still vulnerable to this problem, making it hard to apply to the open world (See Section \ref{subsec:exps_visual_complex}).

%\textbf{Spatial relationships}. 
%We adopt the technique used to construct object relationships from CLEVR and identify four spatial relationships between all pairs of objects: ``left'', ``right'', ``behind'', and ``in front''. 
%Each CATER video contains object coordinates in all video frames.
%In each frame, we can use coordinates of any two objects to compute their relative positions against camera viewpoint vectors projected on the ground plane. 
%However, generating questions that invoke spatial relationships per video frame is computationally expensive and not practical in real world domain. 
%Instead, we compute spatial relationships per certain video intervals. 
%We define two types of video intervals below.

\begin{figure}[t]
	\centering
	\resizebox{1.0\columnwidth}{!} {
	\includegraphics{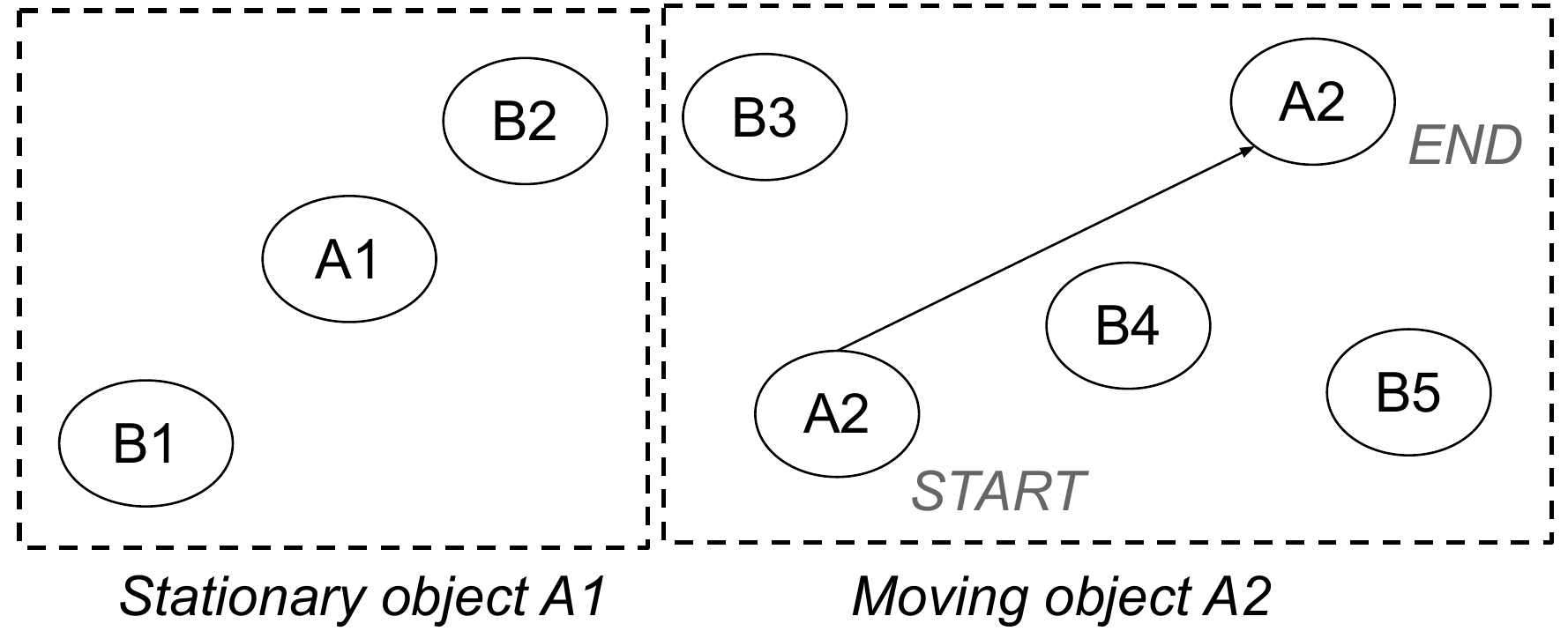}
	}
	\caption{
	%\textbf{Left}: examples of video intervals, defined as continuous time period between two events. Atomic intervals are non-overlapping time periods and all remaining periods are compositional intervals.
	\textbf{Example spatial relationship}: We demonstrate the projection of objects and their movements on the ground plane. Considering the ``left'' relationship, ``A1 is left of B2'' and ``A2 is left of B5''.
	}
	\label{fig:interval_framework}
	\vspace{-0.1in}
\end{figure}

\begin{figure*}[h]
	\centering
	\resizebox{1.0\textwidth}{!} {
	\includegraphics{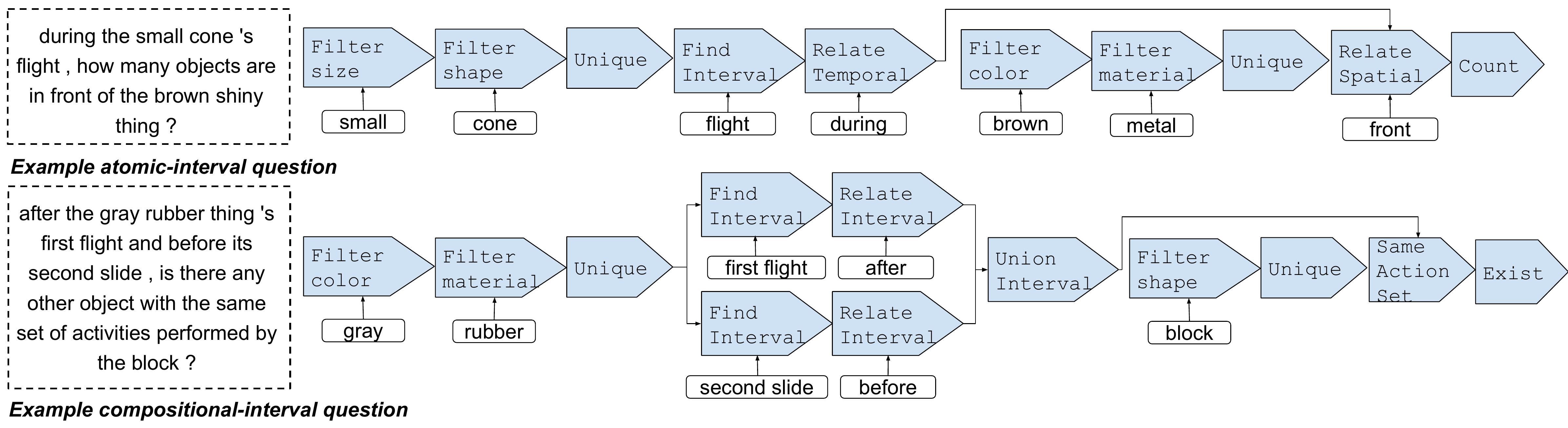}
	}
	\caption{
	\textbf{Example questions and their functional programs:}
	Top: A question of atomic interval with relative spatial relationship.
	Bottom: A question of compositional interval with action set comparison semantic.
	}
	\label{fig:functional_program}
	\vspace{+0.1in}
	\resizebox{1.0\textwidth}{!} {
	\includegraphics{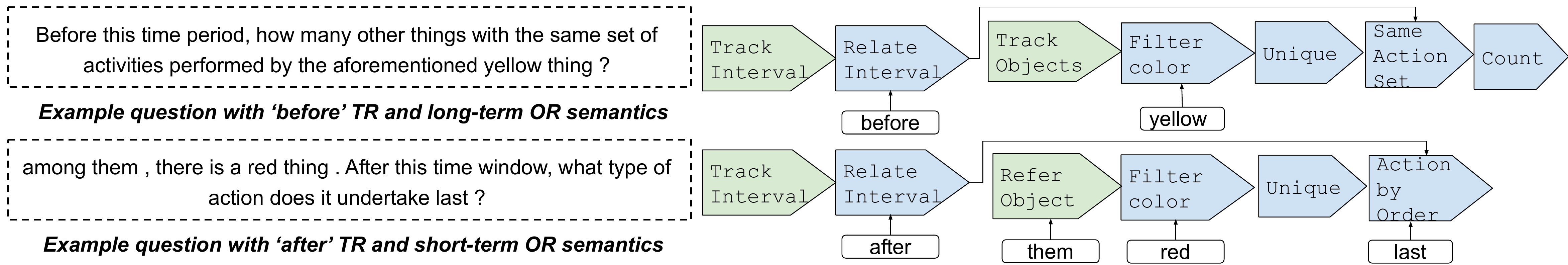}
	}
	\caption{\textbf{Examples questions positioned in dialogue and their functional programs}:
	Each question contained references to past dialogue turns, through video temporal relation (TR) or dialogue object reference (OR). 
	}
	\label{fig:dial_programs}
	\vspace{-0.2in}
\end{figure*}

\noindent \textbf{Video intervals.}
We define video intervals as continuous video frames, limited by a start and end point, each of which can be the start or end of an object's action or the start or end of the whole video. 
We formulate two types of video intervals:

\noindent \textit{1) Atomic intervals.} 
In these intervals, all objects have at most one action and they can be in only one of the two states: in motion or stationary. 
To find atomic intervals, we simply collate the start and end timestamps of all object actions in a CATER video and sort them chronologically. 
By definition, any non-overlapping interval between two timestamps is considered atomic. 
%For example, in Figure \ref{fig:interval_framework}, all time ranges $(t_0,t_1)$, $(t_1,t_2)$, ..., $(t_6, T)$) are atomic. 
This constraint allows us to identify the \emph{relative spatial relationships} (``left'', ``right'', ``behind'', and ``front'') between any two objects by using their coordinates at the start and end of the interval.
Note that in the CATER universe, all actions can be projected either as a straight line (``flying'' and ``sliding'') or a single point (``rotating'' and ``no action'').
Practically, we focus on spatial reasoning only when one of the two objects is stationary.
%We use this object as a ``base'' to compute the relative position of the remaining objects. 
Figure \ref{fig:interval_framework} demonstrates the ``left'' spatial relation, and Figure \ref{fig:functional_program} (Top) shows an example question of atomic interval with spatial relation. 

\noindent \textit{2) Compositional intervals.}
Compositional intervals are all other intervals that are not atomic. In these intervals, an object can have more than one actions, i.e. be in more than one states such as ``flying'' then ``no action''.
Therefore, its movement projections are not linear and we do not identify spatial relations in these cases. 
Instead, we focus on information such as \emph{action set} and \emph{action sequence} to generate questions. 
Figure \ref{fig:functional_program} (Bottom) presents an example question of compositional interval.
%This type of information is derived from object action time interval annotation available in CATER, using simple sorting and lookup methods. 

\noindent To create DVD questions, we first identify all intervals in a video (with a minimum duration of about ~$0.5$s), then randomly sample one interval, and proceed to create questions based on object movements and locations in this interval. 
\noindent Figure \ref{fig:data_distr}-(a) shows the percentages of DVD questions by video interval types. 
Overall, more than $60\%$ of questions are of compositional intervals and among the atomic-interval questions, the majority of them contain a spatial relation. 
We still maintain a small percentage of temporal-agnostic instances (``none'' type) to keep the dialogue flow natural. 

\subsection{Question and Dialogue Generation}
\begin{figure*}[htbp]
	\centering
	\resizebox{1.0\textwidth}{!} {
	\includegraphics{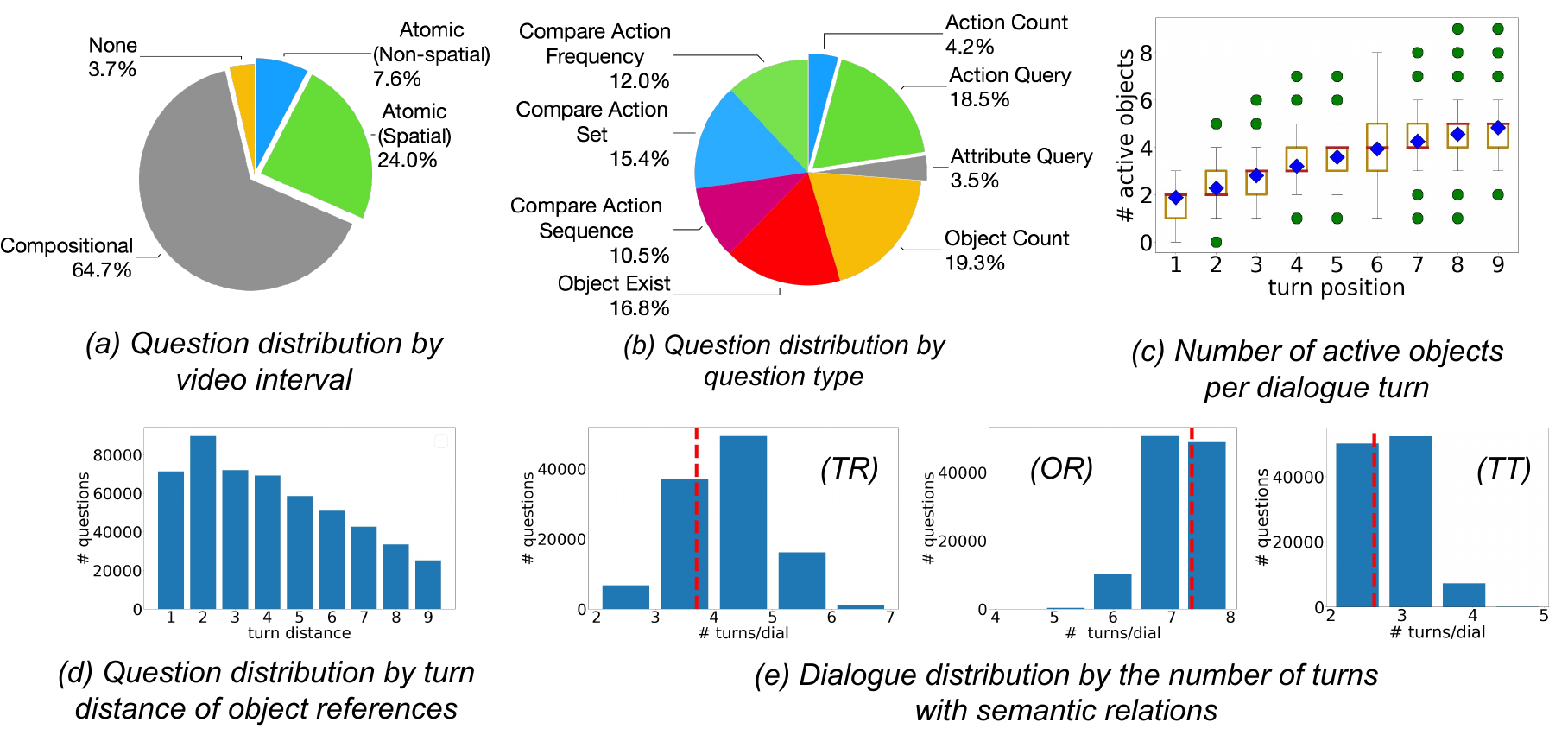}
	}
	%\resizebox{1.0\textwidth}{!} {
	%\includegraphics{PieChart-1.pdf}
	%\includegraphics{PieChart-2.pdf}
	%}
	\caption{
	\textbf{Data analysis of DVD:} 
	(a) and (b): Questions are distributed by 8 question types and 3 video interval types. 
	(c): The boxplot displays the distribution of active objects mentioned in each dialogue turn position. 
	(d): At turn position $i$, an old object originally mentioned in a prior turn position $j$ might be reused, resulting in reference of turn distance $i-j$. 
	(e): Each dialogue turn is incorporated with semantic relations, including TR (temporal relation), OR (object references), and TT (topic transfer). 
	The dotted line indicates the overall average. 
	}
	\label{fig:data_distr}
	\vspace{-0.2in}
\end{figure*}
\noindent \textbf{Question representation.}
We use question templates to materialize questions in natural language. 
Each template associates with an applicable type of video interval and a functional program. 
%As shown in Figure \ref{fig:functional_program}, from CLEVR, we adopt the question families and templates with several extension to video domain. 
Compared to CLEVR functional programs \cite{johnson2017clevr}, we introduce 17 new functional modules, of which 13 are extended for video-based inputs and 4 are extended for dialogue-based inputs. 
%$9$ additional symbolic operations to retrieve information along the temporal aspect of video. 
%Each question is built by compositions of logical operation blocks. 
Overall, we utilize $26$ question templates for 8 question types. 
Figure \ref{fig:functional_program} illustrates two sample questions with corresponding reasoning structures and Figure \ref{fig:data_distr}-(b) shows the statistics of question type distribution. 
Please refer to the supplementary material for the full details of functional modules and question types and examples. 
%One question incorporates spatio-temporal reasoning in an atomic interval while the other contains action set in a compositional interval. 

\begin{figure*}[htbp]
	\centering
	\resizebox{1.0\textwidth}{!} {
	\includegraphics{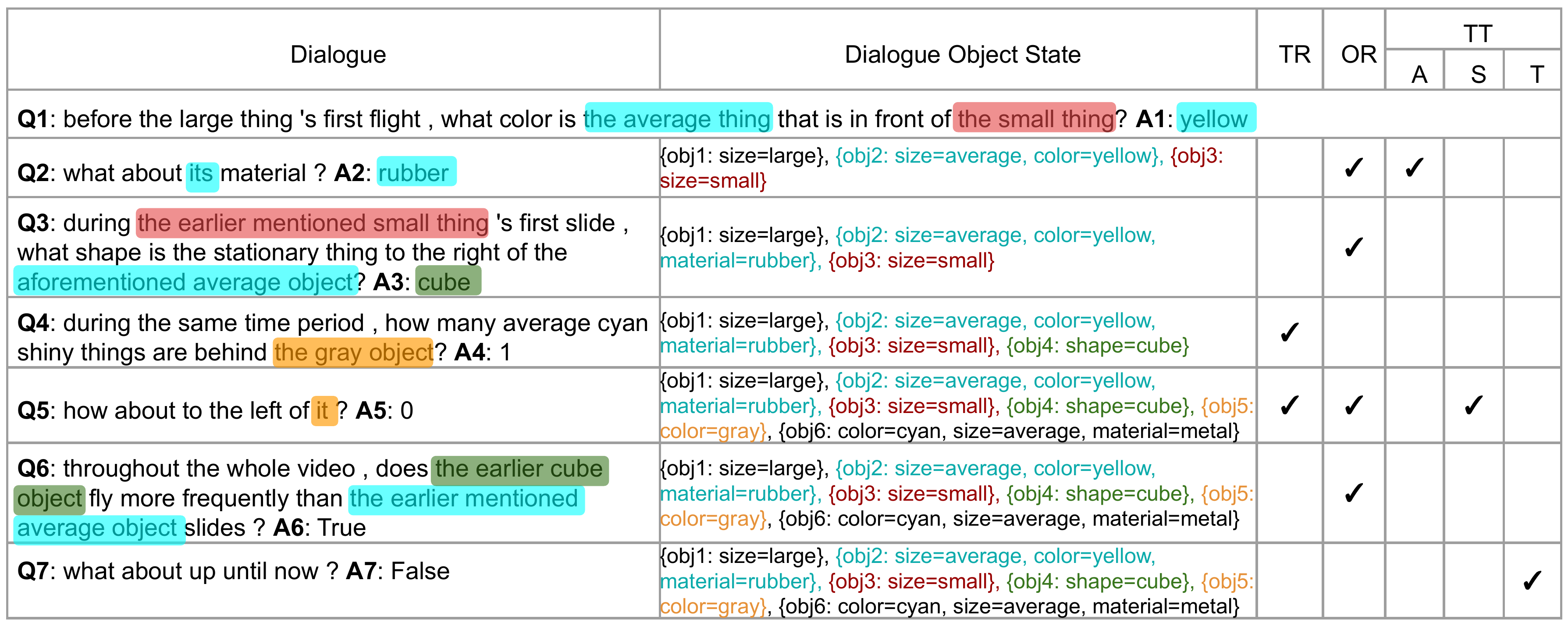}
	}
	\caption{
	\textbf{Dialogue generation:} 
	In each dialogue turn, we generate questions with randomly sampled cross-turn dependencies: temporal relation (TR), object reference (OR), and topic transfers (TT), including attribute (A), spatial (S), and temporal (T) transfer. 
	We maintain a dialogue object state of active objects which are color-coded.
	}
	\label{fig:dial_framework}
	\vspace{-0.2in}
\end{figure*}

\noindent \textbf{Dialogue Generation.}
We generated dialogues with a fixed length of $10$ turns.
In each turn, we adopted a Depth First Search (DFS) approach, as similarly used in CLEVR \cite{johnson2017clevr}, to instantiate questions by sequentially traversing and executing functional programs.
To generate linguistic dependencies between dialogue turns, at each turn, we randomly sample and incorporate one or more of the 3 semantic relations below. 
Figure \ref{fig:dial_programs} and \ref{fig:dial_framework} present examples of 2 questions and a dialogue with these semantic relations.  

\textit{Type I: Video Temporal Relation (TR):} This type of semantic relation tests a system to localize video intervals in relation to past dialogue turns. We randomly select one of three types of relation: (1) \textit{``during''} relation reuses the same time interval as the last dialogue turn, e.g. the Q4 in Figure \ref{fig:dial_framework}; 
(2) \textit{``before''} and (3) \textit{``after''} relations simulate a dialogue flow with references to the earlier and subsequent video segments.
TR synthesizes scenarios when humans either maintain or shift their attention temporally from one video segment to a related part. 

\textit{Type II: Dialogue Object Reference (OR):} We incorporate object references into a question by replacing original object phrase, such as ``the large rubber cone'', with pronouns, such as ``it'', to refer to object(s) mentioned in the earlier part of the dialogue.
The distance of reference is one turn and we call this a \emph{short-term memory} OR. 
Additionally, we simulate \emph{long-term memory} OR by injecting unique objects mentioned further in the past dialogue turns. 
We simulate this behavior by maintaining a \emph{dialogue object state} at each turn.
To choose an object for references, we randomly sample a past dialogue turn position and sample an object introduced in this turn. 
This object then replaces the original object phrases in the question of the current turn. 
For example, in question Q3 in Figure \ref{fig:dial_framework}, ``the earlier mentioned small thing'' is identified from the object originally introduced in Q1.
Following this method, our dialogue simulates scenarios in which humans only focus on a subset of objects rather than all objects in the video scene and they can refer to those objects again over multiple dialogue turns. 
Figure \ref{fig:data_distr}-(c) displays the boxplot of the number of active objects involved in each turn position.
Out of 10 objects (the maximum number of objects in a CATER video), 2 to 5 objects are involved on average per dialogue. 
Figure \ref{fig:data_distr}-(d) shows the question distribution by the turn distance of long-term memory OR, with the majority of questions containing 2-turn distance references.

\textit{Type III: Topic Transfer (TT):} This relation tests the model ability to memorize and reuse the context of the last dialogue turn to the current turn through $3$ types of topic transfers: 
(1) Attribute transfer and (2) spatial transfer reuse the same question from the prior dialogue turn with a modification of object attribute or spatial relation (e.g. Q2 and Q5 in Figure \ref{fig:dial_framework}).
Compared to TR, these two types of topic transfers focus on human attention shifts in spatial space rather than temporal space;
(3) Temporal transfer introduces a unique  setting of \emph{situated dialogue} in DVD. 
Instead of using a fixed video input for each dialogue instance, 
at the first dialogue turn, we shorten a CATER video by a cutoff point, e.g. $T_0$. 
At each later turn, for $30$\% of time, we update the current video input to a new cutoff point later than the previous one e.g. $T_{i+1} > T_i$.
We do not update when the cutoff reaches the end of the original CATER video $\mathcal{T}$ i.e. $T_{i+1}=\mathcal{T}$.
For instance, in Figure \ref{fig:dial_framework}, at Q7, we reuse the same context from Q6 but with new extended visual content.
We introduce temporal transfer as a preliminary step to challenge dialogue systems in a dynamic environment with a continuous visual stream.

\noindent After sampling question templates and semantic dependencies, the ground-truth answers are obtained by executing corresponding functional programs.
For each question template, we discard dominating instances to maintain an approximate uniform distribution of answer values, minimizing bias resulting from question-conditioned data distributions.
Additionally, at each turn, we remove any question that is ill-posed or becomes redundant when positioned in dialogue.
For instance, the question \textit{``how many red rubber objects are there?''} is removed if in a prior dialogue turn, the question is \textit{``how many red objects are there?''} and the answer is already \textit{``1''}.
To do this, we perform a check at every dialogue turn to determine whether involving objects and their attributes are already mentioned in the dialogue object state. 
Finally, we only keep dialogues that have cross-turn dependencies in 9 out of 10 turns, considering the first turn semantically independent. 
Figure \ref{fig:data_distr}-(e) provides the distribution of dialogues by the number of TR, OR, and TT relations. 
For more analysis of DVD, please refer to the supplementary material.

\section{Dialogue Systems on DVD}
\label{sec:exps}
% Chinnad: Discuss the performance of models that only using questions; that we try to make minimal bias that models can rely on
% Ahmad: we want to show that these models are actually required to ground the video information to answer the questions; e.g. discuss performance of cases that do not use video information 
% Performance: Transformer (Q only or H only) 

The video-grounded dialogue task in DVD is defined as a turn-based retrieval task from multiple-choice candidate answers.
%Specifically, a DVD system is given a video $\mathcal{V}$ and dialogue $\mathcal{D}$.
At each dialogue turn $i$ ($i=1,2,...,10$), video input $V_i$,
the ground-truth dialogue context, including question and answer pairs up to the last dialogue turn, $\mathcal{C}_i=(\mathcal{Q}_k, \mathcal{A}_k)|_{k=1}^{k=i-1}$, the question of the current turn $\mathcal{Q}_i$, are provided. 
The system is given a set of candidate answers $\mathcal{A}$, predefined as all possible answer values for all question types, with $|\mathcal{A}|=40$ in DVD, and is required to select one answer from $\mathcal{A}$.
We evaluate models by the accuracy of predicted answers against the ground-truth answers. 
For a system denoted as $\theta$, the objective function is: $ \hat{\mathcal{A}}_i = \argmax_{\mathcal{A}} \displaystyle P(\mathcal{A}_i|\mathcal{V}_i, \mathcal{Q}_i, \mathcal{C}_i; \theta)$.

\subsection{Experimental Setup}
\begin{table*}[htbp]
\centering
\resizebox{1.0\textwidth}{!} {
\begin{tabular}{lccccccccc|c}
\hline
                           Accuracy & \begin{tabular}[c]{@{}c@{}}Answer \\ Prior\end{tabular} & \begin{tabular}[c]{@{}c@{}}Q-type \\ (Random)\end{tabular} & \begin{tabular}[c]{@{}c@{}}Q-type\\ (Freq)\end{tabular} & \begin{tabular}[c]{@{}c@{}}Q-retrieval\\ (TF-IDF)\end{tabular} & \begin{tabular}[c]{@{}c@{}}RNN\\ (Q)\end{tabular} & \begin{tabular}[c]{@{}c@{}}HRNN\\ (C+Q)\end{tabular} & \begin{tabular}[c]{@{}c@{}}HRNN\\ (C+Q)+\\CNN(V)\end{tabular} & \begin{tabular}[c]{@{}c@{}}HRNN\\ (C+Q)+\\ TA(V)\end{tabular} & \begin{tabular}[c]{@{}c@{}}TF\\ (C+Q\\ +V)\end{tabular} & Human \\ \hline
All                    & 21.3                                                    & 27.8                                                       & 35.3                                                    & 32.1                                                           & 39.7                                              & 45.8                                                 & 49.3                                                           & 50.2                                                          & 51.1                                                    & 89.3      \\ 
 Action count               & 0.0                                                     & 9.3                                                        & 23.4                                                    & 19.8                                                           & 16.3                                              & 28.2                                                 & 37.8                                                           & 36.0                                                          & 38.8                                                    & 87.5      \\
Action query               & 0.0                                                     & 12.7                                                       & 23.7                                                    & 20.6                                                           & 25.8                                              & 33.1                                                 & 36.7                                                           & 38.6                                                          & 39.4                                                    & 88.1      \\
 Attribute query                 & 0.0                                                     & 32.9                                                       & 38.7                                                    & 39.4                                                           & 38.1                                              & 39.2                                                 & 43.3                                                           & 45.1                                                          & 43.1                                                    &   98.0    \\
Compare action seq    & 33.4                                                    & 34.1                                                       & 37.3                                                    & 35.1                                                           & 45.5                                              & 52.5                                                 & 58.2                                                           & 57.5                                                          & 61.6                                                    &  91.5     \\
Compare action set         & 25.1                                                    & 28.2                                                       & 36.3                                                    & 28.2                                                           & 32.8                                              & 40.0                                                 & 43.0                                                           & 44.3                                                          & 45.4                                                    &     82.9  \\
Compare action freq                & 48.5                                                    & 50.0                                                       & 50.5                                                    & 44.4                                                           & 58.4                                              & 56.9                                                 & 62.3                                                           & 65.2                                                          & 67.1                                                    &  88.5     \\
Object count                  & 0.0                                                     & 9.1                                                        & 23.3                                                    & 18.8                                                           & 26.2                                              & 38.6                                                 & 40.0                                                           & 40.2                                                          & 39.9                                                    &  90.6     \\
Object exist                  & 48.9                                                    & 49.8                                                       & 51.1                                                    & 54.4                                                           & 66.4                                              & 67.0                                                 & 69.2                                                           & 69.4                                                          & 69.0                                                    &      92.3\\ \hline
None                       & 0.0                                                     & 32.1                                                       & 38.3                                                    & 39.0                                                           & 38.3                                              & 39.5                                                 & 43.1                                                           & 45.1                                                          & 43.4                                                    &     99.1  \\
Atomic (non-spatial)       & 18.8                                                    & 26.3                                                       & 31.9                                                    & 42.4                                                           & 47.2                                              & 47.8                                                 & 49.9                                                           & 50.7                                                          & 48.9                                                    &     83.3  \\
Atomic (spatial)           & 21.2                                                    & 27.3                                                       & 35.5                                                    & 27.6                                                           & 36.8                                              & 46.0                                                 & 47.5                                                           & 47.6                                                          & 47.1                                                    &     93.9  \\
Compositional              & 22.8                                                    & 28.0                                                       & 35.4                                                    & 32.1                                                           & 40.0                                              & 45.8                                                 & 50.2                                                           & 51.4                                                          & 53.2                                                    &     87.1  \\ \hline
Transfer (attribute)       & 0.0                                                     & 30.7                                                       & 45.5                                                    & 37.1                                                           & 40.8                                              & 45.7                                                 & 54.5                                                           & 57.3                                                          & 57.7                                                    &     100.0  \\
Transfer (spatial)         & 49.8                                                    & 42.4                                                       & 44.9                                                    & 26.4                                                           & 29.6                                              & 48.1                                                 & 47.7                                                           & 47.4                                                          & 48.0                                                    &     90.5  \\
Transfer (temporal)        & 28.9                                                    & 38.4                                                       & 22.6                                                    & 3.0                                                            & 30.2                                              & 53.5                                                 & 62.2                                                           & 64.6                                                          & 69.0                                                    &      79.8 \\ \hline
\end{tabular}
}
\caption{
\textbf{Experiment results on the DVD test split}:
Models are evaluated by overall accuracy and by question types (Top), accuracy by video intervals in question (Center), and transferability accuracy (Bottom).
%In addition, they are evaluated by \emph{transferability} metric on question turns with topic transfers, including attribute, spatial, and temporal transfers. 
}
\label{tab:results}
\vspace{-0.2in}
\end{table*}

\textbf{Baselines}. We experimented with a representative set of baseline approaches on DVD, including: 
(1) \emph{Answer Prior}, which selects the most popular answer option as predicted answers;
(2) \emph{Q-type (Random/Frequency)}, which assume known question types and select a random or most popular answer from the corresponding answer space;
(3) \emph{Q-retrieval (TF-IDF)}, which retrieves the most similar question from the training set and use its answer as the predicted answer;
(4) \emph{RNN(Q)} and \emph{HRNN(C+Q)}, which encode dialogue-only components without seeing visual information to predict answers;
(5) \emph{HRNN(C+Q)+CNN(V)/TA(V)}, same as (4) but with access to visual information which is encoded by pretrained CNN models and temporal attention (TA) \cite{jang2017tgif, lei-etal-2018-tvqa, hori2019avsd};
(6) \emph{TF(C+Q+V)}, which uses a Transformer-based architecture to encode visual and language information \cite{schwartz2019factor, le-etal-2019-multimodal, li2020bridging}.
Finally, we conducted internal human evaluation on a subset of the DVD test split. 
For each test sample, a human received an input video, dialogue history, and the question for the current turn. 
The human was required to select an answer from the list of 40 candidates $\mathcal{A}$ to answer the question.
%For more details of baseline models, please refer to the supplementary material. 

\textbf{Experiments}.
Video-grounded dialogues entail a lot of visio-linguistic and reasoning challenges that are not easy to be studied in isolation using existing datasets.
To address this issue with DVD, we exploit the rich annotations of DVD in our experiments during evaluation.
We designed our experiments to systematically analyze model capabilities and shortcomings through unique challenges in video-grounded dialogue systems.
Specifically, in Section \ref{subsec:overall_results}, we analyzed the results of all models overall as well as by each question type. 
In Section \ref{subsec:exps_visual_complex}, we leverage the spatio-temporal annotation of visual objects to analyze model performance by related video interval types, spatial reasoning (results by object containment), and temporal reasoning (results by relative interval length).
In terms of dialogue contextual complexity, in Section \ref{subsec:exps_turns_complex}, we use cross-turn relation annotations to analyze model performance by temporal-based attention shift (TR), dialogue turn distance (OR), and short-term transferability (TT).

\subsection{Results}
\label{subsec:overall_results}
From Table \ref{tab:results} (Top), we observe that ``blind'' systems that use answers only or questions only, achieve quite poor results up to $39\%$ accuracy.
By selecting the most popular answer option, \emph{Answer Prior} only achieves $21\%$ accuracy. 
%For question types with only binary answer options such as ``compare action frequency'' questions, the performance of \emph{RNN(Q)} is not far above results from random guess. 
When a ``blind'' model has access to dialogue history, the performance increases up to $45\%$.
This increment shows that dialogue context contains useful information for a dialogue system to infer answers. 
%However, the performance increase is not significant and it is most likely coming from dialogue turns incorporated with topic-transfer relations. 
We note that on average there are nearly $3$ out of $10$ question turns with a topic transfer per dialogue (see Figure \ref{fig:data_distr}-(e)).
In such cases, a model can randomly make a good guess by just reusing the answer of the last question turn.  
When a system is presented with the visual input, we observe model performance increases up to $51\%$. 
However, in the best system, the performance is still far below the human level with a performance gap of $38$ absolute points. 

In Table \ref{tab:results} (Top), from the results of \emph{Q-type(Random)} per question type, we observed that answers are balanced in each question type. 
The table also shows performance drops between pairs of object-oriented vs. action-oriented question types.
For instance, \emph{TF(C+Q+V)} achieves $38\%$ accuracy in Action count vs. $39\%$ in Object count, and $39\%$ accuracy in Action query vs. $43\%$ in Attribute query. 
In comparison-based questions, comparing action sets tend to be more challenging than comparing action sequences.
To compare action sets of two objects in a video interval, a system needs to process the interval completely. 
However, to compare action sequences, in most cases, the system can determine the answer after the first few action steps the objects perform.
For more analysis of question types and sub-types, please refer to the supplementary material. 

%All existing baselines are not sufficient enough to tackle questions involving both spatial and temporal dependencies, as shown by the very marginal gain in the \emph{atomic(spatial)} results.
%Specifically, we noted that challenging questions with very marginal improvements are ``action count'', ``attr. query'', ``atomic(spatial)'', and ``transfer(spatial)''.
%These questions require strong reasoning ability over both objects' appearance and their movements. 

\subsection{Analysis by Visual Complexity}
\label{subsec:exps_visual_complex}

To understand the drive of the performance by visual inputs, we investigated the results by the visual complexity in questions. 
In Table \ref{tab:results} (Center), compared to \emph{HRNN(C+Q)+CNN(V)}, models using attention, either through \emph{TA(V)} or Transformer, show more improvement in compositional interval questions with increments up to $3$ absolute points. 
%Attention methods such as dot-product attention have been used to select relevant information along the video temporal dimension. 
In other types of intervals, the performance gains are not very significant. 
Particularly, in atomic-interval questions that require spatial localization, the performance does not change when applying attention.
This observation necessitates systems that focus on both spatial and temporal space of visual inputs. 

\begin{figure}[t]
	\centering
	\resizebox{1.0\columnwidth}{!} {
	\includegraphics{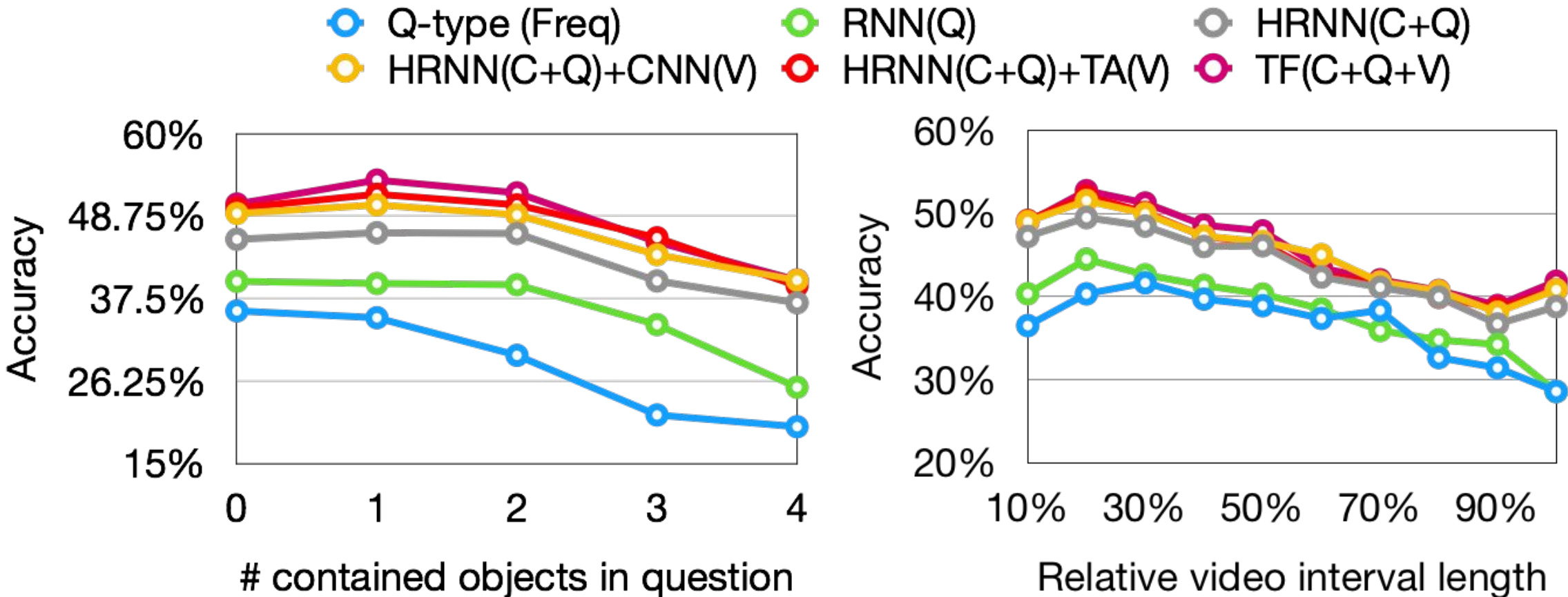}
	}
	\caption{
	\textbf{Experiment results by visual properties}:
	Left: results by the number of objects mentioned in question that are contained in video scenes.
	Right: results by the relative length of video interval in question.
	}
	\label{fig:result_contained_obj}
	\vspace{-0.2in}
\end{figure}

In Figure \ref{fig:result_contained_obj} (Left), we analyzed model performance by the number of objects mentioned in questions that are contained in video scenes. 
We noted that current models are vulnerable to visual object containment, as the accuracy decreases by the number of contained objects. 
This observation is consistent with the results of CATER action recognition tasks \cite{Girdhar2020CATER}.
In Figure \ref{fig:result_contained_obj} (Right), we investigated model performance by the relative length of ground-truth video interval in question, measured as the percentage of the whole video length. 
To make a fair analysis, we removed cases in which a question can be solved correctly without localizing the specific video interval but simply using the whole video.
%and compute the results on the remaining questions. 
%This approach is similar used by \citet{johnson2017clevr} to analyze model performance of spatial localization in CLEVR. 
We observed that model performance decreases as the interval length increases, demonstrating the challenge of long-term video understanding in video scenes. 
We noted that there is a drop in performance in the lowest range of interval lengths, $0-10\%$. 
As this range often represents atomic intervals, the majority of which include questions with spatial relations, systems are negatively affected and the curve drops initially in this low range. 

\subsection{Analysis by Cross-turn Relations}
\label{subsec:exps_turns_complex}
\begin{figure}[t]
	\centering
	\resizebox{1.0\columnwidth}{!} {
	\includegraphics{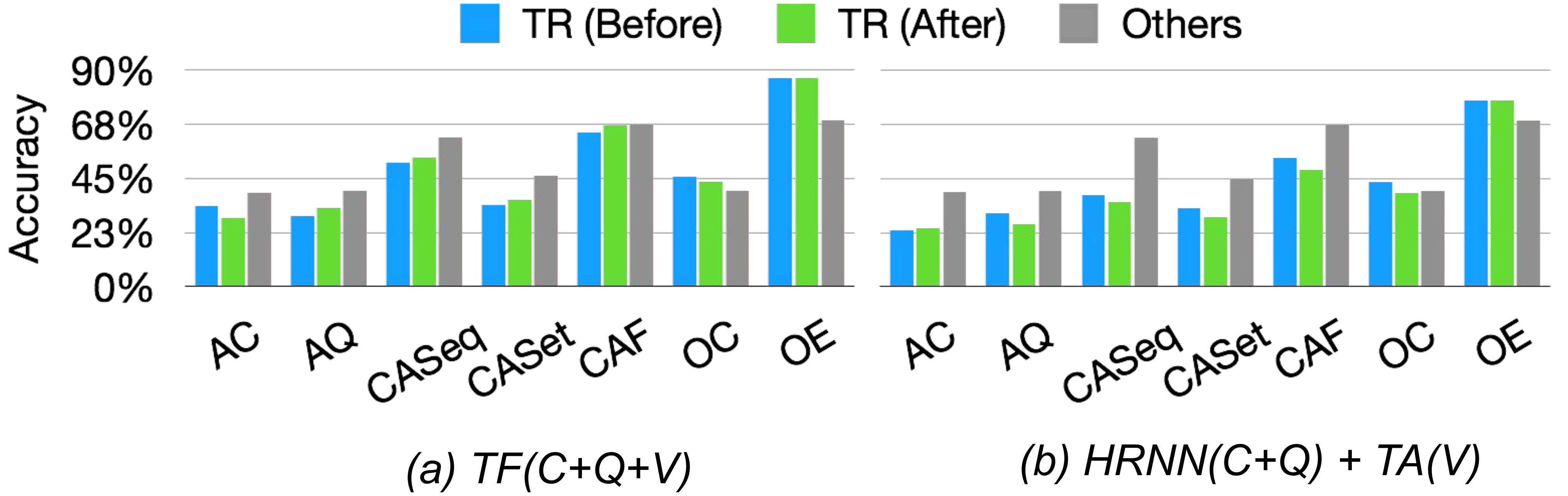}
	}
	\caption{
	\textbf{Experiment results by temporal relations}:
	Action count (AC), Action query (AQ), Attribute query (AttQ), Compare action sequence (CASeq), Compare action set (CASet), Compare action frequence (CAF), Object count (OC), and Object exist (OE).
	}
	\label{fig:result_tr}
	\vspace{-0.1in}
\end{figure}

We examined model performance in a multi-turn setting by cross-turn semantic relations.
First, we investigated the effect of TR. 
In a TR-injected question, a system is required to learn to retrieve a video segment related to the last used segment. 
However, some questions may be correctly answered without localizing the correct segments.
For instance, at the current dialogue turn, a question is of interval $(t_m, t_n)$ and at the next turn, a question with an ``after'' TR is of interval $(t_n, t_q)$ (s.t. $t_m<t_n<t_q$) might be solved if the visual context is the same in both intervals. 
We separate such question turns and measured the results of the remaining questions with TR relations ``after'' and ``before''. 
From Figure \ref{fig:result_tr}, we observed that current systems are not optimal to learn to shift attention to related intervals,
%in relation to the interval used in previous dialogue turns, 
depending on the type of questions. 
In action-based questions (AC, AQ, CASeq, CASet, and CAF), the results of ``before'' and ``after'' TR are lower than those without a TR relation, but in object-based questions (OC, OE), we observed differently. 
This difference can be explained by the dynamics of actions vs. objects.
Between video intervals, information about object actions (e.g. frequency, types) tends to change more easily than objects themselves. 
Action-based questions challenge systems through cross-turn temporal reasoning more than object-based questions.

\begin{figure}[t]
	\centering
	\resizebox{1.0\columnwidth}{!} {
	\includegraphics{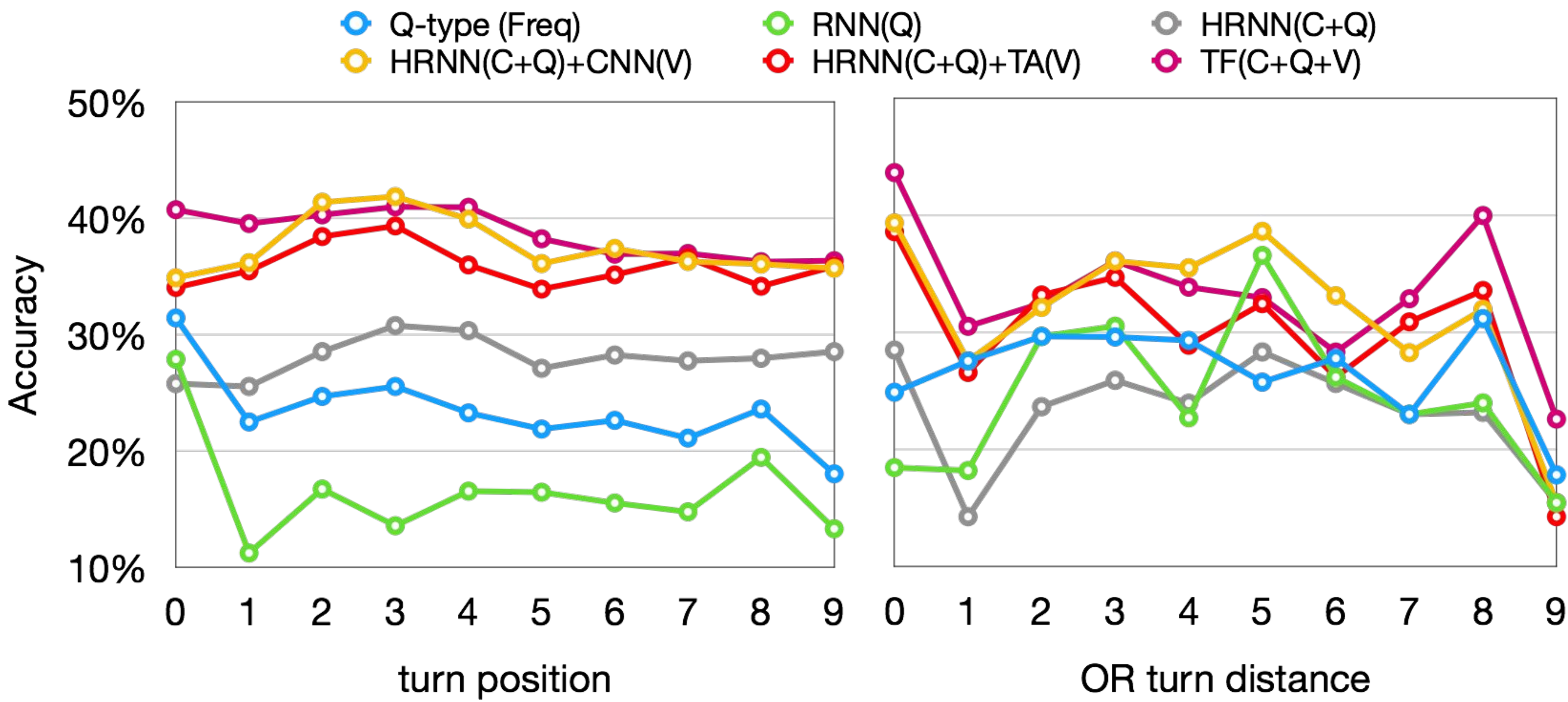}
	}
	\caption{
	\textbf{Experiment results for cross-turn reasoning}:
	Results of Action count questions by turn position (Left) and by turn distance of object references (Right).
	}
	\label{fig:result_or}
	\vspace{-0.1in}
\end{figure}

Secondly, we analyzed the impacts of \emph{long-term memory} OR. 
From Figure \ref{fig:result_or} (Left), we noticed that model performance becomes more stable in systems where dialogue history is introduced as an input. 
For instance, compared to \emph{RNN(Q)}, the performance curve of \emph{TF(C+Q+V)} follows a more gentle downward trend from low to high dialogue turn positions. 
To fairly analyze performance by OR turn distance, we discard any instances that do not require systems to use dialogue context to resolve the references, but simply rely on the input video. 
For example, a question with a reference ``the earlier mentioned red object'' is removed if there is indeed only one ``red object'' in the video scene. 
From results by OR turn distance in Figure \ref{fig:result_or} (Right), we observed all systems are relatively unstable, even as dialogue history is introduced as an input. 
This difference against the results by turn position exhibits a limitation of current systems as they struggle to resolve object references by existing dialogue encoding techniques. 

Finally, to analyze the effect of TT relations, we investigate a new metric, called \emph{transferability}, in Table \ref{tab:results} (Bottom).
When a system is presented with a question turn with a topic transfer, it should learn to derive the answer in relation to the context of the last dialogue turn. 
If the last answer is right, an intelligent system should be able to consistently answer in the current turn correctly. 
For instance, given a question-answer pair ``what is the color of the sliding cube? red'', a human can often infer the answer to a TT(A)-injected question ``what about its material?'' based on the same visual object. 
We gather questions that precede questions containing topic transfers and call this set $\mathcal{Q}^\mathrm{tt}_\mathrm{prior}$.
For each question $q^\mathrm{tt}_\mathrm{prior}$ that the model answered correctly, we measure the accuracy over the corresponding transferred question $q_\mathrm{tt}$ and average the scores.
We observed a clear performance gain from \emph{RNN(Q)} to \emph{HRNN(C+Q)} in terms of transferability metric, demonstrating the impacts of dialogue context on TT questions. 
A chance-based system can achieve approximately $50\%$ transferability by just recycling answers from prior turns.
The best system results, however, are still far from human-level performance.  
This observation necessitates systems designed with a better contextual memory to adapt past context in new dialogue turns. 

\section{Discussion and Conclusion}
We have introduced DVD, a diagnostic dataset designed to analyze video-grounded dialogue systems. 
DVD dataset is generated with tight control of data bias through balancing the question and answer distribution and questions are built based on a principled approach to reflect the complexity in videos and dialogues. 
Our results have shown that DVD can provide interesting insights into system abilities and limitations. 
Specifically, our analysis has revealed some key shortcomings of current models, including: 
(1) limited ability to efficiently integrate visual information from both spatial and temporal space;
(2) limited ability to recognize and compile multiple actions in long-ranged video intervals;
(3) inconsistent performance across dialogue turns, especially in cases when systems are required to switch attention temporally; and
(4) unstable performance to resolve object co-reference in the dialogue context, especially when the turn distance of the object references increases.

These insights provide potential avenues where we hope DVD will be a useful benchmark to explore new ideas.
Specifically, we discuss two research directions:

\textbf{Dialogue object tracking.}
To further diagnose a dialogue system, we aim to study their \emph{long-term memory} reasoning ability to track objects and their attributes mentioned in the dialogue context. 
%Different from topic transfer relations, evaluating system ability to learn from object references is much harder. 
We are inspired by research work of dialogue state tracking in task-oriented dialogues \cite{DBLP:conf/iclr/BordesBW17} and propose to use tracking accuracy metric in video-grounded dialogue systems. 
At each turn $t$, a video-grounded dialogue system should be able to track and update a dialogue state $\mathcal{S}_t$, defined as a set of all mentioned objects $o^t_i$ and their attributes, including sizes $z^t_i$, colors $c^t_i$, materials $m^t_i$, and shapes $s^t_i$: 
$\mathcal{S}_t=(o^t_1, o^t_2,...)=((z^t_1, c^t_1, m^t_1, s^t_1),(z^t_2, c^t_2, m^t_2, s^t_2),...)$.
We define two tracking metrics, including \emph{joint accuracy}, measuring the accuracy of prediction of all objects and attributes as a set, and \emph{slot accuracy}, measuring the accuracy of predicted attributes individually. 
The introduction of these evaluation metrics necessitates a new learning task, dialogue object tracking (DOT) in video-grounded dialogue systems, to better understand current systems' long-term reasoning ability.

\textbf{Video interval tracking.}
Another aspect of dialogue systems that we want to diagnose is their ability to localize video segments in a multi-turn setting. 
Each question turn often focuses on different parts of the video as the dialogue extends over time. 
It is important to learn how a system can localize the right segments of the video from turn to turn. 
Similar to DOT, we define a new learning task for video interval tracking (VIT) in a similar nature as text-to-clip tasks \cite{anne2017localizing}.
The task can be defined as a ranking task of segment candidates to choose the relevant segments in each question turn. 
This task is evaluated by ranking metrics such as Rank$@1$ or Rank$@2$, and \emph{mean intersection over union} (mIoU).  
Alternatively, we can adapt \emph{grounding}, a simple metric used by \citet{Hudson_2019_CVPR} to assess spatial attention of image regions. 
in DVD, \emph{grounding} can be used in temporal attention-based approaches to determine model ability to localize the right position of video intervals in question.

%We hope that the DVD benchmark will facilitate new research ideas and lead to interesting insights to design better systems capable of complex video and dialogue reasoning. 
Finally, we want to emphasize that DVD is designed as a synthetic dataset for diagnosis purposes to systematically evaluate model capabilities.
The benchmark should not be used to replace data of human dialogues but be used to supplement real-world dialogue datasets. 

%\section{Conclusion}
%\label{sec:discussion}
%In this paper, we introduced DVD, a novel diagnostic benchmark to study reasoning capabilities of video-grounded dialogue systems. 
%We described the dataset generation process, provided baseline experiments, and analyzed model abilities and limitations by various aspects. 
%As potential avenues for future work, we consider two sub-tasks to improve dialogue systems: 
%(1) Video interval tracking (VIT) task requires a system to explicitly identify which video segment each dialogue turn is referring to.
%This task is an extension of prior research in temporal localization through text, or text-to-clip \cite{anne2017localizing}, but is designed in a multi-turn setting;
%(2) Dialogue object tracking (DOT) task has the dialogue state tracking (DST) nature, which requires a system to track information slots in task-oriented dialogues \cite{mrksic-etal-2017-neural}. 
%While in task-oriented dialogues, the tracked slots are used to create API queries to entity databases, tracked objects in DOT is used to solve object references and locate the visual objects from videos.

\bibliographystyle{acl_natbib}
\bibliography{anthology,acl2021}

\appendix

\section{A Comparison of DVD to Related Benchmarks}
 \begin{table*}[htbp]
\resizebox{1.0\textwidth}{!} {
%\begin{tabular}{@{\extracolsep{4pt}}p{10cm}ccccc@{}}
\begin{tabular}{p{10.5cm}ccccc}
\hline 
\multicolumn{1}{c}{\multirow{2}{*}{Benchmarks}} 
&  \multirow{2}{*}{\begin{tabular}[c]{@{}c@{}}Diagnostic\\benchmark\end{tabular}} 
& \multicolumn{2}{c}{Visual reasoning}
& \multicolumn{2}{c}{Language reasoning}\\
\cline{3-4}
\cline{5-6}
\multicolumn{1}{c}{}                          
& \multicolumn{1}{c}{}
& SR    
& TR
& DOT
& VIT\\
\hline
{\textbf{Image/video QA, embodied QA}} & & & & &\\
VQA \cite{antol2015vqa}, Visual7W \cite{zhu2016visual7w} & \xmark & \cmark & \xmark & \xmark & \xmark\\
TGIF-QA \cite{jang2017tgif}, TV-QA \cite{lei-etal-2018-tvqa}                             & \xmark                                                                                & \cmark                     & \cmark                    &   \xmark                   &   \xmark                   \\
IQA \cite{gordon2018iqa}, EQA  \cite{eqa_matterport}                                         & \xmark                                                                                & \cmark                     & \cmark                    &   \xmark                   &   \xmark                   \\
\hline
\textbf{Image/video grounded dialogues, navigation dialogues} \\
VisDial \cite{das2017visual}, GuessWhat \cite{de2017guesswhat}                       &   \xmark                                                                              & \cmark                     &  \xmark                    & \cmark                    & \xmark                     \\
AVSD \cite{hori2019avsd}, CVDN \cite{thomason:corl19}                                           & \xmark                                                                                & \cmark                     & \cmark                    & \cmark                    & \cmark                    \\
\hline
\textbf{Synthetic image/video QA} \\
% \multicolumn{7}{l}{\textbf{Synthetic image/video QA}} \\
SHAPE \cite{andreas2016neural}, CLEVR \cite{johnson2017clevr}                                           & \cmark                                                                               & \cmark                     &  \xmark                    &  \xmark                    &   \xmark                   \\
SVQA \cite{song2018explore}, CLEVRER \cite{Yi*2020CLEVRER}                                                     & \cmark                                                                               & \cmark                     & \cmark                    & \xmark                     &   \xmark                   \\\hline
\textbf{Synthetic dialogues} \\
bAbI \cite{DBLP:conf/iclr/BordesBW17}                              & \cmark                                                                               & \xmark                     &\xmark                      & \cmark                    &  \xmark                    \\
MNIST Dialog \cite{seo2017visual}, CLEVR-Dialog \cite{kottur-etal-2019-clevr}                              & \cmark                                                                               & \cmark                     &\xmark                      & \cmark                    &  \xmark                    \\
\hline \hline
\textbf{DVD (Ours)}                                    & \cmark                                                                               & \cmark                     & \cmark                    & \cmark                    & \cmark                   \\
\hline
\end{tabular}
}
\caption{
\textbf{Comparison to related benchmarks:}
Compared to existing datasets for vision-language understanding,
DVD is the first diagnostic benchmark designed for both spatial reasoning (SR) and temporal reasoning (TR) and 
explicit requiring dialogue object tracking (DOT) and video interval tracking (VIT) in a multi-turn setting.
}
\label{tab:related_work}
\end{table*}

In Table \ref{tab:related_work}, we compare DVD with related benchmarks by 4 aspects: spatial reasoning (SR), temporal reasoning (TR), dialogue object tracking (DOT), and video interval tracking (VIT). 
SR and TR are visual-related reasoning types.
SR refers to the reasoning requirement to localize information within an image. 
SR is the most popular reasoning type, being involved in most vision-language benchmarks such as VQA \cite{antol2015vqa} and TGIF-QA \cite{jang2017tgif}. 
TR is often present when a video is used as input, which requires systems to localize the relevant temporal location in the video.  
However, TR is not just limited to video understanding tasks but also refers to problems with dynamic visual inputs such as navigation systems or embodied QA. 
DOT and VIT refer to cross-turn semantic relations in a multi-turn dialogue problem setting. 
DOT refers to the use of object references, requiring systems to learn to resolve these references in dialogue context.  
DOT can be seen clearly in most dialogue benchmarks as object references are used frequently in traditional dialogues. 
VIT is a new reasoning requirement in video-grounded dialogue tasks. 
It requires systems to localize temporal parts of the video from turn to turn. 
VIT is less obvious in prior benchmarks as it is challenging to simulate.
It is mostly present in specific tasks such as AVSD \cite{hori2019avsd} and CVDN \cite{thomason:corl19} where a video input is introduced and at each turn, only a specific temporal part of the video is relevant. 
Compared to existing benchmarks, DVD is the first diagnostic benchmark that combines all 4 aspects, SR, TR, DOT, and VIT, together. 

\section{DVD Functional Program Modules}
In Table \ref{tab:data_type} and \ref{tab:functional_prog}, we describe all data types and functional program modules in DVD. 
In total, there are $20$ data types and $32$ functional modules. 
Among the functional modules, compared to CLEVR \cite{johnson2017clevr}, we introduced 17 novel modules that are designed to be executed on dialogue or video components.
Within these modules, there are 13 video-based modules (\emph{Count Action, Filter Action, Same Action Set, Same Action Sequence, Find Interval, Union Interval, Relate Spatial, Relate Temporal, Query Action Set, Query Action Sequence, Action by Frequency, Action by Order, Equal Action}) and 4 dialogue-based modules (\emph{Refer Object, Track Object, Refer Interval, Track Interval}). 

\begin{table*}[htbp]
\centering
\resizebox{0.9\textwidth}{!} {
\begin{tabular}{lp{13cm}}
\hline
Data type         & Description                                                                                                                      \\ \hline
\textit{Object}            & A dictionary storing the attributes of an object, including its shape, size, color, and material, and details of its actions, including start and end points                                                                               \\
\textit{Objects}           & A list of of \textit{Objects}                                                                                                                \\
\hline
\textit{Spatial Relation}  & A value from the set: ``left'', ``right'', ``front'', and ``behind''                                                                                    \\
\textit{Temporal Relation} & A value from the set: ``before'', ``after'', and ``during''                                                                                         \\
\hline
\textit{Reference}         & Pronoun, such as ``it'', ``its'', ``them'', ``the first one'', used to refer to an object or action mentioned in the last dialogue turn                                                           \\
\textit{Last Turn}    & The last dialogue turn, including the last question and answer                                                                            \\
\textit{Object Tracker}    & A list of \textit{objects}, storing all objects involved and their attributes mentioned so far up to the last dialogue turn                                       \\
\textit{Interval Tracker}  & A list of video intervals mentioned so far up to the last dialogue turn                                                                             \\
\hline
\textit{Interval}          & A tuple containing the start and end time of a video segment                                                                    \\
\textit{Action}            & Any value from ``sliding'', ``flying'', ``rotating'', and ``no action''                                                                  \\
\textit{Action Set}        & Any combination of \textit{actions}, except for ``no ation'', without duplication. A standalone ``no action'' is acceptable.                                                                                    \\
\textit{Action Sequence}   & Any combination of \textit{actions}, except for ``no action'', that can form a sequence. A standalone ``no action'' is acceptable.                                                                                     \\
\textit{Frequency}   & A positive integer that indicates the number of times an action is undertaken. Frequency can also be expressed by superlatives such as ``least'' or ``most''.                                                                                      \\
\textit{Order}   & An ordinal number that indicates the order of an action during a video interval.                                                                                      \\
\hline
\textit{Color}             & A string that indicates an object color: ``gold'', ``gray'', ``green'', ``purple'', ``red'', ``cyan'', ``cylinder'', ``blue'', ``brown'', ``yellow'' \\
\textit{Material}          & A string that indicates an object material, including ``metal'' and ``rubber''                                                       \\
\textit{Shape}             & A string that indicates an object shape, including ``cone'', ``cube'', ``sphere'', ``snitch''                                            \\
\textit{Size}              & A string that indicates an object size, including ``large'', ``medium'', and ``small''                                                 \\
\hline
\textit{Binary}            & A binary value, either ``False'' or ``True''                                                                                         \\
\textit{Integer}           & An integer value \textgreater{}= 0                                                                                          \\
\hline
\end{tabular}
}
\caption{
\textbf{Data types in DVD:} In total, there are $20$ data types, which can be categorized by the following groups (from Top to Bottom): object-based, relation-based, cross-turn based, action-based, attributes, and binary/integer.
}
\label{tab:data_type}
\end{table*}

% \usepackage{multirow}
\begin{table*}[htbp]
\resizebox{1.0\textwidth}{!} {
\begin{tabular}{p{1.4cm}|p{3cm}p{3cm}p{1.4cm}p{7cm}}
\hline
\begin{tabular}[c]{@{}c@{}}Module \\ Type\end{tabular}                     & \multicolumn{1}{c}{\begin{tabular}[c]{@{}c@{}}Module \\ Name\end{tabular}} & \multicolumn{1}{c}{\begin{tabular}[c]{@{}c@{}}Input \\ Type\end{tabular}}            & \multicolumn{1}{c}{\begin{tabular}[c]{@{}c@{}}Output \\ Type\end{tabular}} & \multicolumn{1}{c}{Module Description}                                             \\
\hline

\multirow{2}{*}{Count}                                                     & \texttt{Count Object}                                                              & \textit{Objects}                                                                              & \textit{Integer}                                                                    & Return number of objects                                       \\
\cline{2-5}
                                                                           & \texttt{Count Action}                                                               & \textit{(Interval, Object, Action)}                                                                  & \textit{Integer}                                                                    & Return number of times an object undertakes a specific type of actions during an interval             \\
                                                                           \hline
Exist                                                                      & \texttt{Exist}                                                                      & \textit{Objects}                                                                              & \textit{Binary}                                                                     & Return whether there is at least one resulting object from the last module         \\
\hline
\multirow{9}{*}{\begin{tabular}[c]{@{}c@{}}Object-\\ based\end{tabular}}   & \texttt{Filter Color}                                                               & \textit{(Objects, Color)}                                                                     & \textit{Objects}                                                                    & Return objects of a specific color                                                 \\
\cline{2-5}
                                                                           & \texttt{Filter Material}                                                            & \textit{(Objects, Material)}                                                                  & \textit{Objects}                                                                    & Return objects of a specific material                                              \\
                                                                           \cline{2-5}
                                                                           & \texttt{Filter Shape}                                                               & \textit{(Objects, Shape)}                                                                     & \textit{Objects}                                                                    & Return objects of a specific shape                                                 \\
                                                                           \cline{2-5}
                                                                           & \texttt{Filter Size}                                                                & \textit{(Objects, Size)}                                                                      & \textit{Objects}                                                                    & Return objects of a specific size                                                  \\
                                                                           \cline{2-5}
                                                                           & \texttt{Filter Action}                                                              & \textit{(Interval, Objects, Action)}                                                                    & \textit{Objects}                                                                    & Return objects performing a specific action during an interval                                        \\
                                                                           \cline{2-5}
                                                                           & \texttt{Same Action Set}                 & \textit{(Interval, Object)}                                                                               & \textit{Objects}                                                                    & Return objects performing the same action set as another object during an interval                    \\
                                                                           \cline{2-5}
                                                                           & \texttt{Same Action Sequence}          & \textit{(Interval, Object)}                                                                               & \textit{Objects}                                                                    & Return objects performing the same action sequence as another object during an interval               \\
                                                                           \cline{2-5}
                                                                           & \texttt{Unique}                                                                     & \textit{Objects}                                                                              & \textit{Object}                                                                     & Return the unique object from resulting objects                                    \\
                                                                           \cline{2-5}
                                                                                                   & \texttt{Scene}                                                                     &                                                                               & \textit{Objects}                                                                     & Return all objects in the current video                                    \\

                                                               \hline
\multirow{2}{*}{\begin{tabular}[c]{@{}c@{}}Interval-\\ based\end{tabular}} & \texttt{Find Interval}                                                              & \textit{(Object,  Action)}                                                                    & \textit{Interval}                                                                   & Return the start and end point of the interval of an action performed by an object \\
\cline{2-5}
                                                                           & \texttt{Union Interval}                                                             & \textit{(Interval$_1$, Interval$_2$)}                                                                 & \textit{Interval}                                                                   & Return the overlapping interval from Interval$_1$ and Interval$_2$                                \\
                                                                           \hline
\multirow{2}{*}{Relate}                                                    & \texttt{Relate Spatial}                                                             &\textit{(Interval, Object, Spatial Relation)}                & Objects                                                                    & Return objects located in relation to another object during a specific interval                               \\
\cline{2-5}
                                                                           & \texttt{Relate Temporal}                                                            & \textit{(Interval, Temporal Relation)}                                                                             & \textit{Interval}                                                                   & Return interval in relation to another interval                                    \\
                                                                           \hline
\multirow{3}{*}{\begin{tabular}[c]{@{}c@{}}Integer\\ -based\end{tabular}}  & \texttt{Greater Than}                                                               & \textit{(Integer$_1$, Integer$_2$)}                                                                 & \textit{Binary}                                                                     & Return whether Integer$_1$ $>$ Integer$_2$                                                                                   \\
\cline{2-5}
                                                                           & \texttt{Less Than}                                                                  & \textit{(Integer$_1$, Integer$_2$)}                                                                   & \textit{Binary}                                                                     & Return whether Integer$_1$ $<$ Integer$_2$                                                                                   \\
                                                                           \cline{2-5}
                                                                           & \texttt{Equal}                                                                      & \textit{(Integer$_1$, Integer$_2$)}                                                                   & \textit{Binary}                                                                     & Return whether Integer$_1$ $=$ Integer$_2$                                                                                    \\
                                                                           \hline
\multirow{4}{*}{Multi-turn}                                                & \texttt{Refer Object}                                                               &\textit{(Reference, Last Turn)}               & \textit{Objects}                                                                    & Resolve object reference based on the last dialogue turn                                                                                   \\
\cline{2-5}
                                                                           & \texttt{Track Object}                                                               & \textit{Object Tracker}                                                                       & \textit{Objects}                                                                    & Return all objects mentioned so far in the dialogue                                                                                   \\
                                                                           \cline{2-5}
                                                                           & \texttt{Refer Interval}                                                             & \textit{(Reference, Last Turn)}               & \textit{Interval}                                                                  & Resolve interval reference to an action mentioned in the last dialogue turn                                                                                    \\
                                                                           \cline{2-5}
                                                                           & \texttt{Track Interval}                                                             & \textit{Interval Tracker}                                                                     & \textit{Interval}                                                                   &   Return the interval used in the last dialogue turn                                                                                  \\
                                                                           \hline
\multirow{5}{*}{\begin{tabular}[c]{@{}c@{}}Action\\ -based\end{tabular}}   & \texttt{Query Action Set}                & \textit{(Interval, Object)}                                                                               & \textit{Action Set}                      &  Return the set of actions performed by an object during an interval                                                                                  \\
\cline{2-5}
                                                                           & \texttt{Query Action Sequence}           & \textit{(Interval, Object)}                                                                               & \textit{Action Sequence}                 &   Return the sequence of actions performed by an object during an interval                                                                                 \\
                                                                           \cline{2-5}
                                                                           & \texttt{Action by Frequency}            & \textit{(Interval, Object, Frequency)}                                                                    & \textit{Action Set}                      &  Return the set of action performed by an object for a fixed number of times during an interval                                                                                  \\
                                                                           \cline{2-5}
                                                                           & \texttt{Action by Order}                 & \textit{(Interval, Object, Order)}                                                                    & \textit{Action}                                                                     & Return an specific action performed by an object during an interval in an ordinal position (e.g. $1^{st}$, $2^{nd}$)                                                                                   \\
                                                                           \cline{2-5}
                                                                           & \texttt{Equal Action}                                                               & \textit{(Action Set/Sequence, Action Set/Sequence)}                                                                   & \textit{Binary}
                                                                           
                                      & Return whether two set of actions are the same or two sequences of actions are the same                                     \\
                                                                           \hline
\multirow{4}{*}{\begin{tabular}[c]{@{}c@{}}Other \\ Modules\end{tabular}}  & \texttt{Query Color}                                                                & \textit{Object}                                                                               & \textit{Color}                                                                      & Obtain the color of a specific object                                                                                   \\
\cline{2-5}
                                                                           & \texttt{Query Material}                                                             & \textit{Object}                                                                               & \textit{Material}                                                                   & Obtain the material of a specific object                                                                                   \\
                                                                           \cline{2-5}
                                                                           & \texttt{Query Shape}                                                                & \textit{Object}                                                                               & \textit{Shape}                                                                      & Obtain the shape of a specific object                                                                                    \\
                                                                           \cline{2-5}
                                                                           & \texttt{Query Size}                                                                 & \textit{Object}                                                                               & \textit{Size}                                                                       & Obtain the size of a specific object              \\
                                                                           \hline
\end{tabular}
}
\caption{
\textbf{Details of functional program modules:}
In total, there are 32 functional program modules, of which 17 are modules introduced for video-based and dialogue-based components. 
}
\label{tab:functional_prog}
\end{table*}

\begin{table*}[htbp]
\resizebox{1.0\textwidth}{!} {
\begin{tabular}{p{1.5cm}|c|c|p{11cm}}
\hline
{  \begin{tabular}[c]{@{}c@{}}Question\\ Type\end{tabular}}                                 & {  \begin{tabular}[c]{@{}c@{}}Question\\ Interval\end{tabular}} & {  \begin{tabular}[c]{@{}c@{}}Question \\ Subtype\end{tabular}} & \multicolumn{1}{c}{{  Example}}                                                                                                                        \\
\hline
                                                                                       &                                                             &  more                                                        & until the end of the snitch 's rotation , does the blue thing fly more frequently than the purple object flies ?                                   \\
                                                                                       \cline{3-4}
{  }                                                                                        & {  }                                                            & {  equal}                                                       & {  during the whole video , does the sphere rotate as frequently as the cylinder slides ?}                                                             \\
\cline{3-4}
\multirow{-3}{*}{{ \begin{tabular}[c]{@{}c@{}}Compare \\ action \\ frequency\end{tabular}}}                                                       & \multirow{-3}{*}{{  Compositional}}                             & {  less}                                                        & {  after the large thing 's first flight , does the cylinder fly less frequently than the green object slides ?}                                       \\
\hline
{  }                                                                                        & {  }                                                            & {  count}                                                       & {  before the large matte thing 's flight , how many other things perform the same sequence of activities as the cyan object ?}                        \\
\cline{3-4}
\multirow{-2}{*}{{  \begin{tabular}[c]{@{}c@{}}Compare \\ action \\ sequence\end{tabular}}} & \multirow{-2}{*}{{  Compositional}}                             & {  exist}                                                       & {  until the end of the metal sphere 's slide , is there any other thing with the same sequence of activities performed by the average purple thing ?} \\
\hline
{  }                                                                                        & {  }                                                            & {  count}                                                       & {  throughout the whole video , how many other things undertake the same types of actions as the large block ?}                                        \\
\cline{3-4}
\multirow{-2}{*}{{  \begin{tabular}[c]{@{}c@{}}Compare \\ action \\ set\end{tabular}}}      & \multirow{-2}{*}{{  Compositional}}                             & {  exist}                                                       & {  until the end of the cyan shiny thing 's last slide , is there any other object that has the same types of actions as the large rubber object ?}    \\
\hline
{  }                                                                                        & {  }                                                            & {  by frequency}                                                & {  during the gray thing 's flight , what activities that the big thing perform the least ?}                                                           \\
\cline{3-4}
{  }                                                                                        & {  }                                                            & {  by order}                                                    & {  until the end of the average green thing 's second flight , what is the purple thing doing first ?}                                                 \\
\cline{3-4}
{  }                                                                                        & \multirow{-3}{*}{{  Compositional}}                             & {all actions}                                                   & {  during the whole video , what is the brown thing doing ?}                                                                                           \\
\cline{2-4}
             & {Atomic (Non-Spatial)}                                                      & {all actions}                                                      & {after the red cube 's second slide , what actions does the  green sphere undertake ?}                \\
\cline{2-4}
\multirow{-5}{*}{{  \begin{tabular}[c]{@{}c@{}}Action \\ query\end{tabular}}}               & {Atomic (Spatial)}                                                      & {all actions}                                                      & {  during the small thing 's second rotation , what actions does the average rubber thing that is in front of the red thing undertake ?}                \\
\hline
{  }                                                                                        & {  }                                                            & {  size}                                                        & {  how big is the cylinder ?}                                                                                                                          \\
\cline{3-4}
{  }                                                                                        & {  }                                                            & {  color}                                                       & {  what color is the cylinder ?}                                                                                                                       \\
\cline{3-4}
{  }                                                                                        & {  }                                                            & {  material}                                                    & {  what material is the brown cone ?}                                                                                                                  \\
\cline{3-4}
\multirow{-4}{*}{{  \begin{tabular}[c]{@{}c@{}}Attribute \\ query\end{tabular}}}            & \multirow{-4}{*}{{  None}}                                      & {  shape}                                                       & {  what is the shape of the cyan object ?}                                                                                                             \\
\hline
Action count                                 & {Compositional}                                               & -                                                           & {throughout the whole video , how many times does the metal cylinder spin in total ?}                                                                \\
\hline
{  }                                                                                        & {  Compositional}                                               & {  -}                                                           & {  throughout the whole video , what number of sliding matte cones are there ?}                                                                        \\
\cline{2-4}
{  }                                                                                        & {Atomic (Spatial)}                                                            & {-}                                                     & {  during the cylinder 's first rotation , what number of objects are in front of the purple thing ?}                                                  \\
\cline{2-4}
{  }                                                                                        &Atomic (Non-spatial)                                    & {-}                                                 & {  before the cylinder 's slide , how many stationary metallic objects are there ?}                                                                    \\
\cline{2-4}
\multirow{-4}{*}{{  \begin{tabular}[c]{@{}c@{}}Object \\ count\end{tabular}}}               & {  None}                                                        & {  -}                                                           & {  what number of purple things are there ?}                                                                                                           \\
\hline
{  }                                                                                        & {  Compositional}                                               & {  -}                                                           & {  throughout the whole video , is there any sliding large rubber cone ?}                                                                              \\
\cline{2-4}
{  }                                                                                        & {Atomic (Spatial)}                                                            & {-}                                                     & {  during the brown thing 's rotation , is there any cone in front of the purple cube ?}                                                               \\
\cline{2-4}
\multirow{-3}{*}{{  \begin{tabular}[c]{@{}c@{}}Object \\ exist\end{tabular}}}               &Atomic (Non-spatial)                                    & {-}                                                 & {  since the start of the big red thing 's flight , is there a contained small red metal cylinder ?}                         \\
\hline
\end{tabular}
}
\caption{
\textbf{Question types and examples:}
In total, there are 8 question types, each of which is designed for one or more types of video intervals (Atomic, Compositional, or None). 
In each question type, we also classify further into question sub-types. 
}
\label{tab:question_types}
\end{table*}

\begin{figure*}[htbp]
	\centering
	\resizebox{0.6\textwidth}{!} {
	\includegraphics{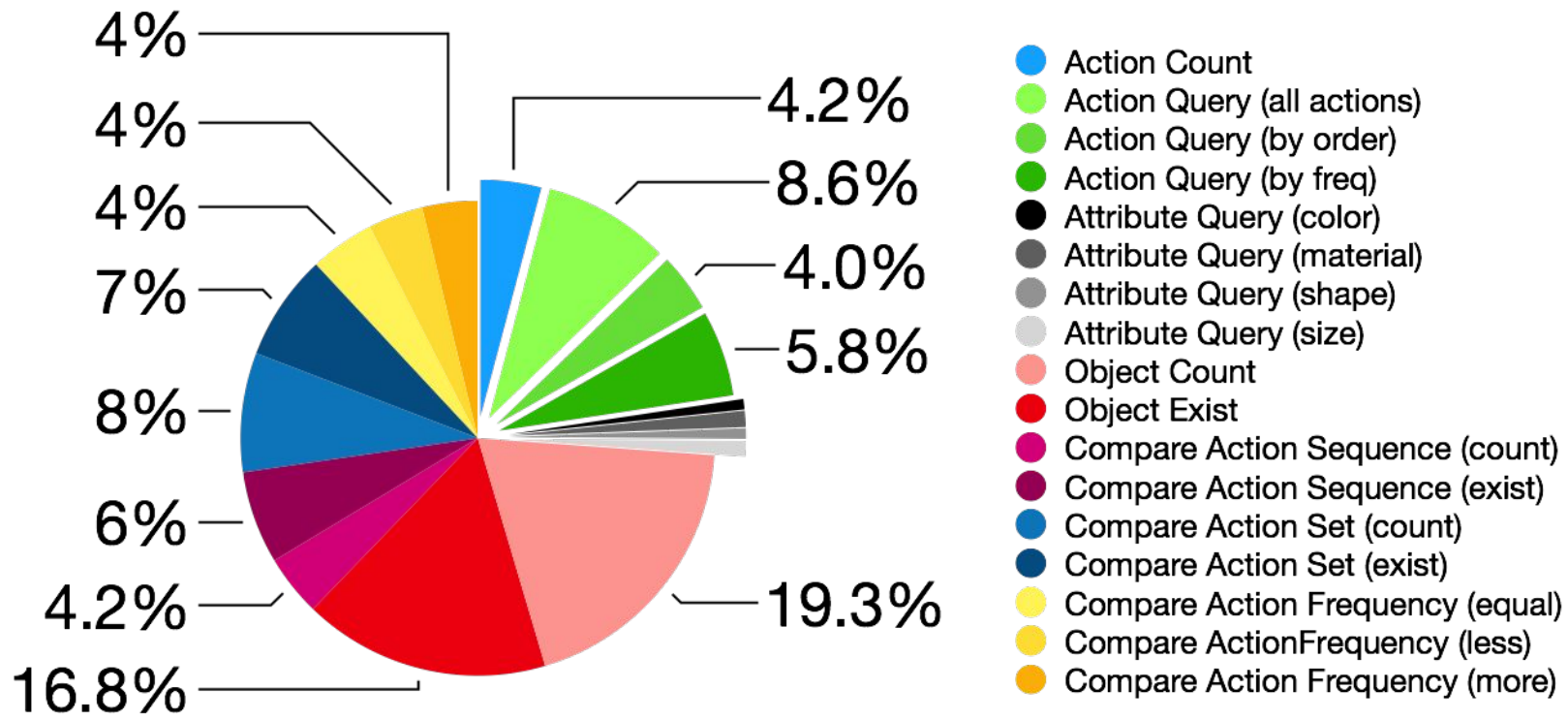}
	}
	\caption{
	\textbf{Distribution of questions by question sub-types:}
	For each question type, we classify questions further into corresponding sub-types. In total, from 8 question types, there are 17 question sub-types. 
	}
	\label{fig:q_subtype_dist}
\end{figure*}

\begin{figure*}[htbp]
	\centering
	\resizebox{0.9\textwidth}{!} {
	\includegraphics{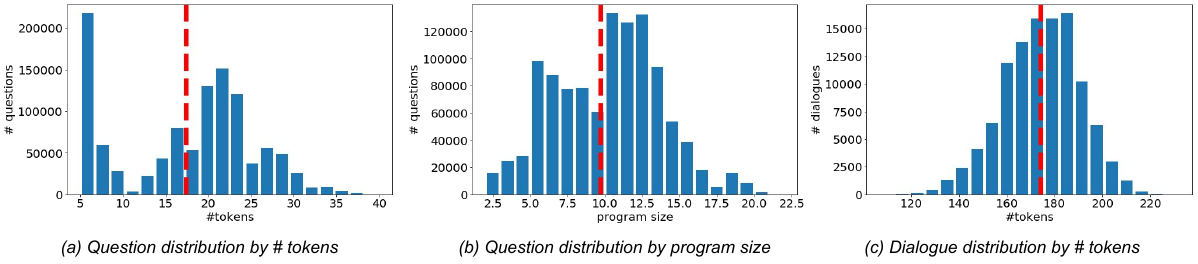}
	}
	\caption{
	\textbf{Distribution of dialogues and questions:}
	Questions and dialogues in DVD are balanced by the program size and text length.
	The dotted line indicates the position of the overall average.
	}
	\label{fig:dial_len_dist}
\end{figure*}

\section{DVD Question Types, Sub-types, and Examples}
In Table \ref{tab:question_types}, we detail all 8 question types for DVD. 
In each question type, we described the types of video intervals applicable, including \emph{Atomic} interval, \emph{Compositional} interval, or \emph{None}.
\emph{None} type is used in temporal agnostic questions, such as questions to query object attributes or count objects. 
In each question type, we further classify questions by question sub-types.
Figure \ref{fig:q_subtype_dist} presents the distribution of questions by question sub-types. 
We observed that per each question type, question sub-types are balanced in most cases. 
For instance, the question type \emph{Compare Action Frequency} include 3 sub-types: \emph{equal}, \emph{less}, and \emph{more}, and each is about $4\%$ of the total questions. 
Similar observations can be seen in other question types, including \emph{Compare Action Sequence}, \emph{Compare Action Set}, and \emph{Attribute Query}. 

\section{Additional DVD Data Statistics}
\subsection{Question and dialogue size}
Figure \ref{fig:dial_len_dist} displays the distributions of questions and dialogues by the number of tokens and by program size. 
By the number of tokens, questions have an average size of more than $17$ tokens. 
There is a spike of questions in the lower range of $5-10$ tokens as this range represents questions with \emph{Topic Transfers} (TT). 
Questions with TT are usually shorter than average, as these questions recycle the information of the preceding questions in the last dialogue turn. 
For instance, a TT-based question with a spatial transfer, ``how about its left ?'', has a length of $5$ tokens.
At the dialogue level, the distribution is more balanced by the number of tokens, with an average length of more than $170$ tokens. 
When measured by functional program size, defined as the number of modules involved, questions are relatively balanced, with an average size of 10 reasoning steps. 

Table \ref{tab:question_type_size} details the number of tokens and program size per question type. 
Shorter question types are query-based questions such as \emph{Action query} and \emph{Attribute query}. 
In \emph{Attribute query}, the questions are shorter in both number of tokens and program size as these questions do not include functional program for temporal localization. 
More complex question types are questions with comparison semantics, including \emph{Compare action freq, Compare action seq}, and \emph{Compare action set}. 

\begin{table}[htbp]
\centering
\resizebox{0.8\columnwidth}{!} {
\begin{tabular}{lcc}
\hline
\multicolumn{1}{c}{Question Type} & \# tokens & Program size \\
\hline
Object exist                      & 15.7      & 11.4         \\
Obj count                         & 15.3      & 10.1         \\
Action count                      & 15.3      & 6.7          \\
Action query                      & 17.4      & 7.7          \\
Attribute query                   & 8.1       & 3.5          \\
Compare action freq                   & 19.7      & 13.4         \\
Compare action seq           & 21.9      & 9.9          \\
Compare action set                & 19.9      & 9.1     \\
\hline
\end{tabular}
}
\caption{
\textbf{Average size per question type:}
Query-related question types such as ``attr query'' and ``action query'' tend to have smaller program size.
Questions requiring comparison such as ``compare int'' and ``compare action'' tend to have larger program size. 
}
\label{tab:question_type_size}
\end{table}

\subsection{Video Distribution}

\begin{figure}[htbp]
	\centering
	\resizebox{0.7\columnwidth}{!} {
	\includegraphics{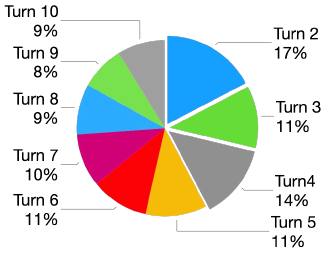}
	}
	\caption{
	\textbf{Distribution of turn positions where video input is updated:}
	To simulate \emph{Topic Transfer (Temporal)} questions, we update the video input from the prior dialogue turn with an additional subsequent segment. 
	}
	\label{fig:video_cutoff}
\end{figure}

As described in the main paper, to simulate the \emph{Topic Transfer (Temporal)}, instead of using a fixed video input throughout all dialogue turns, we simulate a dynamic video input stream. 
At each dialogue turn, for $30\%$ of time, we extend the video input by adding a new video segment at the end of the video. 
The process stops when we reach the original end of a CATER video input. 
At this point, the video input will not be updated until the end of the dialogue. 
Using this method, we observed that each DVD dialogue contains at least $1$ turn and maximum 3 turns in which the video input is updated. 
Figure \ref{fig:video_cutoff} illustrates the distribution of turn positions where video input is updated. 
In general, the video input is updated across all turn positions, with the most likely position is in the $2^{nd}$ dialogue turn and the least likely are in the $9^{th}$ and $10^{th}$ turn. 

\begin{figure*}[htbp]
	\centering
	\resizebox{1.0\textwidth}{!} {
	\includegraphics{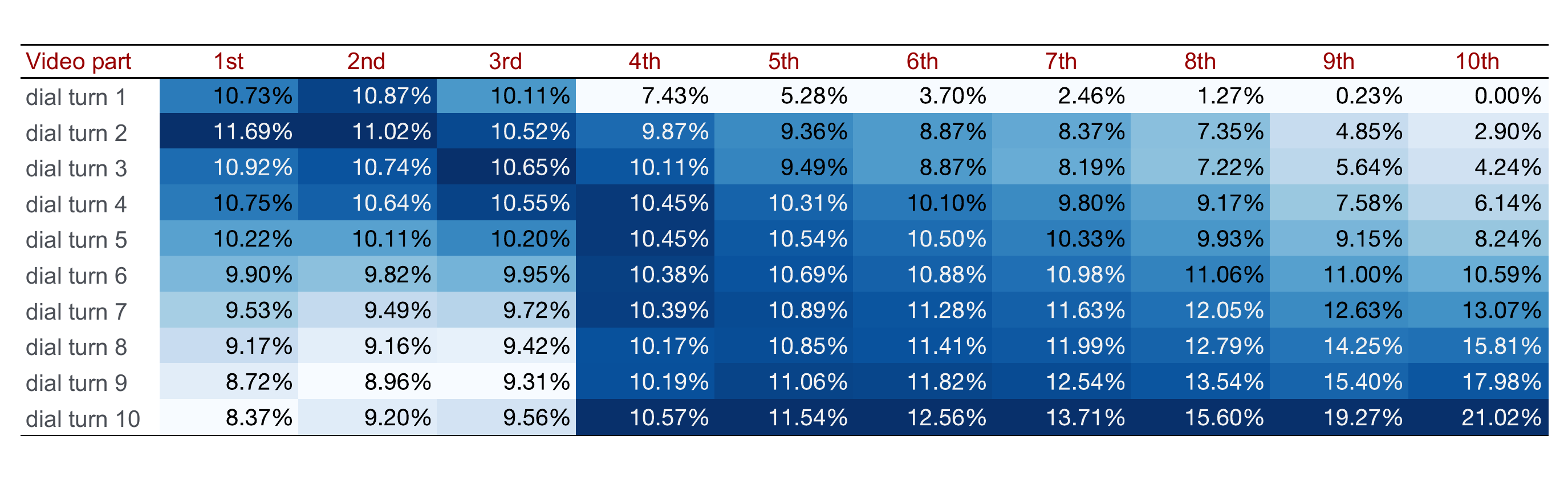}
	}
	\caption{
	\textbf{Distribution of dialogue turn positions where video segments are involved:}
	Our approach to simulate situated dialogues results in more balanced usage of video segments throughout the entire dialogue. 
	In DVD, the earlier video segments tend to be involved in the earlier turns of the dialogue. 
	Likewise, later video segments tend to be mentioned more in the later turns of the dialogue. 
	}
	\label{fig:active_video}
\end{figure*}

Figure \ref{fig:active_video} displays the distribution of dialogue turn positions for each of active video part. 
To identify active video parts, we first divide each original CATER video into 10 equal segments. 
For each dialogue turn, we identify the video interval the question is based on, and find all video parts that overlap with this interval. 
These video parts are considered as active parts for that turn. 
From Figure \ref{fig:active_video}, we observe that active video parts following a diagonal trend: the earlier video parts tend to be mentioned in earlier turn positions, while the later video parts are more active in the later turn position. 
This observation shows balanced attention of the temporal space in DVD question turns, whereby the dialogue flow generally follows a chronological order of video segments. 

\section{Additional Experimental Details}
\subsection{Experimental Procedure Details} 
We used a 3D version of ResNet-101  \cite{hara2018can} pretrained on the Kinetics dataset \cite{kay2017kinetics}.
We extracted all video features from the final average pooling layer, giving $2048$-dimensional features that are not fine-tuned.
The videos were resized to $112\times112$ before feature extraction, and video clips were sampled with a size of 16 frames and striding of 4 frames. 
All LSTM networks used 2 recurrent layers and all MLP networks used ReLU activation with dropout \cite{srivastava2014dropout} and one hidden layer.  
All models were optimized using cross-entropy objective loss between ground-truth and predicted answers with Adam optimizer \cite{DBLP:journals/corr/KingmaB14}. 
We tuned model hyper-parameters using the validation set and selected the best models with the highest accuracy metric to evaluate on the test set. 

\subsection{Baseline Model Details}
\textbf{Answer Prior.} Each answer option is encoded using a token-level LSTM \cite{hochreiter1997long} and scored by a multi-layer perceptron (MLP). This model is trained to select the most popular answer options from the training set without looking at either videos or dialogues. 

\textbf{Q-type.} This baseline selects a random answer (\emph{Q-type Random}) or the most popular answer (\emph{Q-type Frequency}) for each question type. The ground-truth question type is given in this baseline. 

\textbf{Q-retrieval.} At test time, for each question, this model simply computes the cosine similarity to all questions in the training set based on TF-IDF features. The answer to the most similar question is directly chosen as the predicted answer. 

\textbf{RNN(Q).} Question $\mathcal{Q}$ is processed with learned word embeddings and encoded by a toke-level LSTM. 
The final hidden state is passed to an MLP with softmax scores to predict a distribution over answer candidates.
This baseline tests the question-conditional bias.

\textbf{HRNN(C+Q).} When using dialogue context, this model uses a hierarchical architecture with 2 LSTMs to encode dialogue by turn-level and token-level sequence \cite{serban2017hrnn}. The final hidden state of the turn-level LSTM is input to an MLP.  

\begin{table*}[htbp]
\centering
\resizebox{1.0\textwidth}{!} {
\begin{tabular}{lccccccccc}
\hline
                           Model & \begin{tabular}[c]{@{}c@{}}Answer \\ Prior\end{tabular} & \begin{tabular}[c]{@{}c@{}}Q-type \\ (Random)\end{tabular} & \begin{tabular}[c]{@{}c@{}}Q-type\\ (Freq)\end{tabular} & \begin{tabular}[c]{@{}c@{}}Q-retrieval\\ (TF-IDF)\end{tabular} & \begin{tabular}[c]{@{}c@{}}RNN\\ (Q)\end{tabular} & \begin{tabular}[c]{@{}c@{}}HRNN\\ (C+Q)\end{tabular} & \begin{tabular}[c]{@{}c@{}}HRNN\\ (C+Q)+\\CNN(V)\end{tabular} & \begin{tabular}[c]{@{}c@{}}HRNN\\ (C+Q)+\\ TA(V)\end{tabular} & \begin{tabular}[c]{@{}c@{}}TF\\ (C+Q\\ +V)\end{tabular}  \\ \hline
 Action count               & 0.0                                                     & 9.3                                                        & 23.4                                                    & 19.8                                                           & 16.3                                              & 28.2                                                 & 37.8                                                           & 36.0                                                          & 38.8                                                        \\
\rowcolor[HTML]{DCD8D8} Action query               & 0.0                                                     & 12.7                                                       & 23.7                                                    & 20.6                                                           & 25.8                                              & 33.1                                                 & 36.7                                                           & 38.6                                                          & 39.4                                                         \\
\rowcolor[HTML]{DCD8D8}$\hookrightarrow$ All actions & 0.0                                                     & 12.6                                                       & 18.8                                                    & 17.3                                                           & 20.1                                              & 27.0                                                 & 31.0                                                           & 32.8                                                          & 33.1                                                          \\
\rowcolor[HTML]{DCD8D8}$\hookrightarrow$ By order    & 0.0                                                     & 12.7                                                       & 30.4                                                    & 28.9                                                           & 40.2                                              & 41.4                                                 & 45.0                                                           & 48.4                                                          & 49.5                                                           \\
\rowcolor[HTML]{DCD8D8}$\hookrightarrow$ By frequency      & 0.0                                                     & 12.9                                                       & 26.4                                                    & 19.7                                                           & 24.4                                              & 36.4                                                 & 39.4                                                           & 40.4                                                          & 41.8                                                           \\
 Attribute query                 & 0.0                                                     & 32.9                                                       & 38.7                                                    & 39.4                                                           & 38.1                                              & 39.2                                                 & 43.3                                                           & 45.1                                                          & 43.1                                                         \\
$\hookrightarrow$ Color         & 0.0                                                     & 11.0                                                       & 14.7                                                    & 16.4                                                           & 14.5                                              & 17.4                                                 & 19.6                                                           & 22.4                                                          & 21.3                                                          \\
$\hookrightarrow$ Material      & 0.0                                                     & 52.0                                                       & 53.9                                                    & 52.4                                                           & 51.9                                              & 54.6                                                 & 54.6                                                           & 56.6                                                          & 54.2                                                           \\
$\hookrightarrow$ Shape          & 0.0                                                     & 20.4                                                       & 40.3                                                    & 45.2                                                           & 39.9                                              & 37.6                                                 & 46.9                                                           & 49.9                                                          & 47.4                                                           \\

$\hookrightarrow$ Size         & 0.0                                                     & 35.6                                                       & 35.8                                                    & 35.2                                                           & 36.5                                              & 37.3                                                 & 42.9                                                           & 43.1                                                          & 41.0                                                           \\
\rowcolor[HTML]{DCD8D8}Compare action seq    & 33.4                                                    & 34.1                                                       & 37.3                                                    & 35.1                                                           & 45.5                                              & 52.5                                                 & 58.2                                                           & 57.5                                                          & 61.6                                                           \\
\rowcolor[HTML]{DCD8D8}$\hookrightarrow$ Count & 0.0                                                     & 9.4                                                        & 21.0                                                    & 14.7                                                           & 16.1                                              & 33.6                                                 & 35.3                                                           & 36.3                                                          & 38.1                                                           \\
\rowcolor[HTML]{DCD8D8}$\hookrightarrow$ Exist & 55.4                                                    & 50.5                                                       & 48.1                                                    & 48.6                                                           & 64.9                                              & 65.1                                                 & 73.4                                                           & 71.6                                                          & 77.1                                                         \\
Compare action set         & 25.1                                                    & 28.2                                                       & 36.3                                                    & 28.2                                                           & 32.8                                              & 40.0                                                 & 43.0                                                           & 44.3                                                          & 45.4                                                          \\
$\hookrightarrow$ Count & 0.0                                                     & 8.9                                                        & 26.2                                                    & 15.5                                                           & 15.3                                              & 25.7                                                 & 27.3                                                           & 29.4                                                          & 30.4                                                          \\
$\hookrightarrow$ Exist & 53.1                                                    & 49.7                                                       & 47.6                                                    & 42.4                                                           & 52.3                                              & 56.0                                                 & 60.6                                                           & 61.0                                                          & 62.2                                                          \\
\rowcolor[HTML]{DCD8D8}Compare action freq                & 48.5                                                    & 50.0                                                       & 50.5                                                    & 44.4                                                           & 58.4                                              & 56.9                                                 & 62.3                                                           & 65.2                                                          & 67.1                                                          \\
\rowcolor[HTML]{DCD8D8}$\hookrightarrow$ Equal        & 52.2                                                    & 50.0                                                       & 52.2                                                    & 43.0                                                           & 60.8                                              & 59.9                                                 & 65.4                                                           & 65.9                                                          & 68.1                                                           \\
\rowcolor[HTML]{DCD8D8}$\hookrightarrow$ Less         & 45.9                                                    & 49.5                                                       & 45.9                                                    & 44.9                                                           & 58.6                                              & 56.3                                                 & 60.5                                                           & 64.4                                                          & 66.7                                                          \\
\rowcolor[HTML]{DCD8D8}$\hookrightarrow$ More         & 46.8                                                    & 50.4                                                       & 53.2                                                    & 45.6                                                           & 55.5                                              & 54.1                                                 & 60.5                                                           & 65.3                                                          & 66.4                                                         \\
Object count                  & 0.0                                                     & 9.1                                                        & 23.3                                                    & 18.8                                                           & 26.2                                              & 38.6                                                 & 40.0                                                           & 40.2                                                          & 39.9                                                          \\
\rowcolor[HTML]{DCD8D8}Object exist                  & 48.9                                                    & 49.8                                                       & 51.1                                                    & 54.4                                                           & 66.4                                              & 67.0                                                 & 69.2                                                           & 69.4                                                          & 69.0                                                       \\ \hline
\end{tabular}
}
\caption{
\textbf{Experiment results per question type and sub-type on the DVD test split}:
Models are evaluated for accuracy per question type and sub-type.
For each question type, the results of sub-types are detailed in the next rows and marked with a $\hookrightarrow$ symbol. 
}
\label{tab:result_subtypes}
\end{table*}

\textbf{HRNN(C+Q)+CNN(V)/TA(V).} In video-grounded dialogue systems, a video $\mathcal{V}$ is typically represented with features from a pretrained 3D CNN. 
A video is separated into shorter segments/clips, each of which is passed through a CNN model, resulting in temporal-variant features. 
To aggregate video features, we experiment with 2 approaches: 1) \emph{CNN} simply using the CNN video features averaged along with the temporal steps and 2) temporal attention (\emph{TA}) using an attention mechanism to select the relevant temporal steps as similarly adopted by \citet{hori2019avsd}.
The prior text-only systems are integrated with these aggregation methods by concatenating the final text and video representations before passing them to the MLP. 

\textbf{TF(C+Q+V).} Similar to the work of \citet{le-etal-2019-multimodal, schwartz2019factor, li2020bridging}, this model adopts deep attention networks to model cross-modal interactions. We concatenate all text and visual input components as a single sequence and pass them to a Transformer encoder. 
A special ``[CLS]'' token is used in the first position of the sequence to aggregate information through all attention rounds. 
Its final representation is then passed to an MLP to predict answers.

\subsection{Analysis by Question Types and Sub-types}
\begin{figure*}[htbp]
	\centering
	\resizebox{1.0\textwidth}{!} {
	\includegraphics{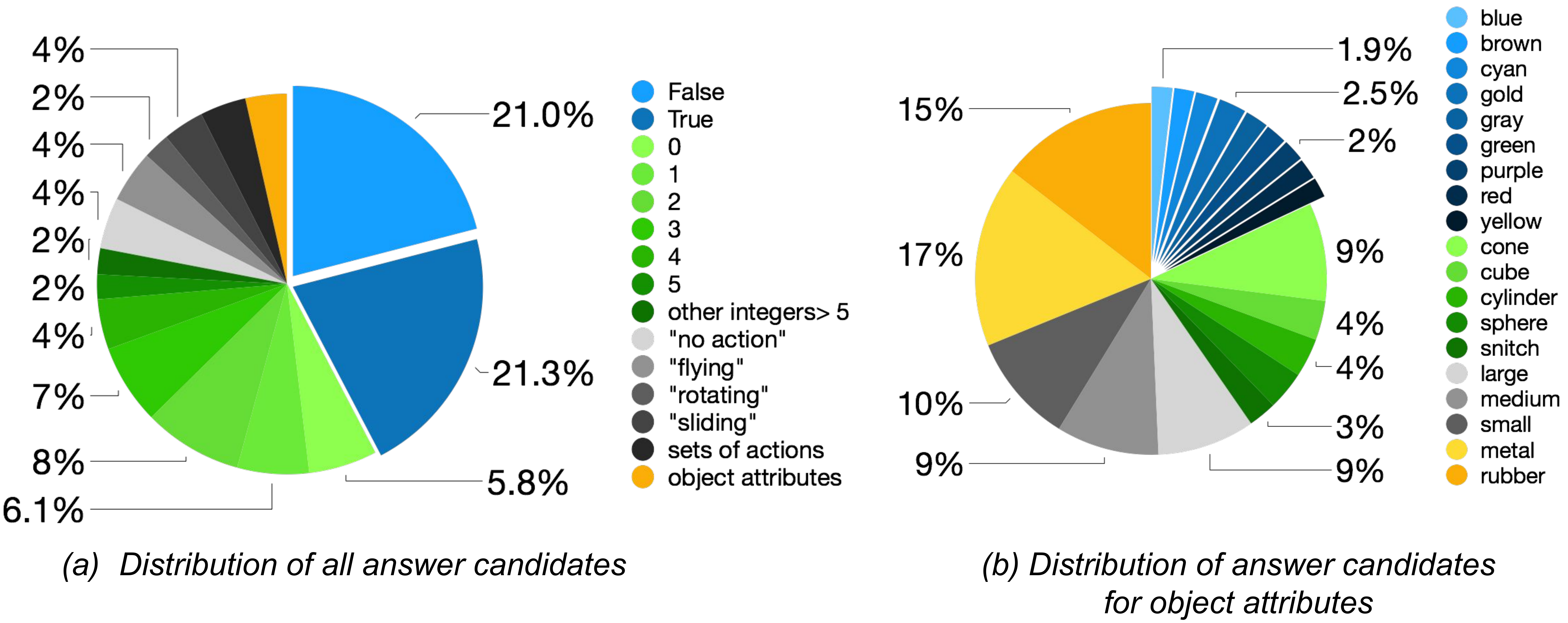}
	}
	\caption{
	\textbf{Distribution of answer options in DVD:} 
	(a) Distribution of all answer candidates, including binary types, integer types, action types, and object attributes. 
	For presentation purposes, rare answer options are grouped, including answers of integers that are more than 5, answers for action sets, and answers for object attributes. 
	(b) More detailed distribution of answer options for object attributes, including options of object colors, shapes, sizes, and materials. 
	}
	\label{fig:ans_distr}
\end{figure*}

Table \ref{tab:result_subtypes} details the experiment results by question types and sub-types. 
To analyze how DVD is controlled to minimize question-conditional bias, especially in ``blind'' models, we present the distribution of answer options in Figure \ref{fig:ans_distr}.
We describe and discuss our findings below. 

\textbf{Action count:} 
These questions ask about the number of times an object performs a specific type of action, e.g. ``how many times does the red cube rotate?''. 
In CATER, there is no limit to the number of times an object undertakes an action, but typically, an object performs an action less than 5 times in a CATER video (See Figure \ref{fig:ans_distr}-(a)).
In these questions, \emph{Q-type(Freq)} and \emph{HRNN(C+Q)} achieves performance of $23-28\%$, showing the DVD questions are controlled by question-conditional bias. 
In systems that use video inputs, the accuracy on this question type increased substantially to $37-38\%$, showing the impacts of visual features. 
However, temporal attention-based models do not perform much better than models without an attention mechanism, with a performance increase of up to $2$ absolute points. 

\textbf{Action query:}
These questions ask about action, either as a set (\emph{all actions}) or a single action (\emph{by order} or \emph{by frequency}). 
There are 3 types of actions in CATER.
We note that the action ``flying'' is more popular than other answer options. 
This observation is probably caused by the design of CATER videos focusing on object containment. 
An object contains another object by flying on top and cover that object completely. 
Therefore, ``blind'' systems will be biased towards this answer option, achieving up to $33\%$ accuracy. 
By question sub-types, overall, we noticed questions asking about action sets is the most challenging since the correct answers can be more than a single action type. Between action queried \emph{by order} and \emph{by frequency}, the latter is more challenging as systems will need to process through an interval completely to learn the action frequency. 

\textbf{Attribute query:}
In the CATER universe, there are 3 object sizes, 2 materials, 5 shapes, and 9 colors. 
\emph{Q-type} model performance in question sub-types are generally closed to the values of $33\%$, $50\%$, and $11\%$ accuracy for question queries of size, material, and colors respectively. 
The performance of \emph{Q-type} on queries of object shape is significantly higher than $20\%$ because there is one dominating answer option, ``cone'' (See Figure \ref{fig:ans_distr}-(i)). 
This behavior is again due to CATER videos focusing on object containment. 
In CATER videos, only objects that have the cone shape can contain another object.
Overall, in attribute query questions, the performance increases as visual input is introduced. 
However, even as attention approaches are used, the performance gains are very marginal. 
This raises a concern for current approaches to learn more fine-grain representations of objects in both spatial and temporal space, including their visual appearance and properties. 

\textbf{Object count/exist:}
In the CATER universe, there is no fixed limit on the number of objects in a video. 
By simply selecting the most answer options, \emph{Q-type (Freq)} achieves performance close to $23\%$ for \emph{Object count} questions and close to a chance-based performance of $50\%$ for \emph{Object exist} questions. 
In \emph{Object exist}, compared to other question types, we noted that there is a substantial performance increase from \emph{Q-type} models to \emph{RNN(Q)}, close to $16\%$ absolution point increase.
This increment shows the benefit of processing questions sequentially token by token. 
Sequentially encoding questions through LSTM and optimzing against the ground-truth answers might result in learning of data-specific constraints in the CATER universe. 
For instance, in CATER videos, object actions are designed to be constrained by object shapes. 
A sphere object cannot rotate in CATER and a cone is often subject to the ``flying'' action to simulate object containment. 

\textbf{Compare action sequence/set:}
These questions are similar to \emph{Object count} and \emph{Object exist} questions,  but with additional semantics that require systems to compare the actions of two objects. 
In both question types, \emph{Q-type} models achieve performance about $21-26\%$ for \emph{Count} questions, and $47-48\%$ for \emph{Exist} questions. 
These findings show that the answer options are balanced within each question sub-type. 
We also observe that the action set-based questions are consistently more challenging than sequence-based questions at the question sub-type level.
For instance, the best model \emph{TF(C+Q+V)} performs better in both \emph{Count} and \emph{Exist} sequence-comparison questions than the corresponding set-comparison questions.

\textbf{Compare action frequency:}
All ``blind'' systems achieve accuracy results within $49-60\%$, not much better than a chance-based model. 
Within the 3 sub-types, questions that simply asking if two objects have \emph{equal action} frequencies are slightly easier to solve than questions about \emph{larger} and \emph{smaller} relations.

\end{document}